\documentclass[10pt]{article} % For LaTeX2e
\usepackage[preprint]{tmlr}
\usepackage{caption}
\usepackage{subcaption}
\usepackage{amsmath}               % great math stuff
\usepackage{amssymb}               % great math symbols
\usepackage{amsfonts}              % for blackboard bold, etc
\usepackage{amsthm}                % for theorems, http://tex.stackexchange.com/a/130655
\usepackage{multirow}
\usepackage{tikz}                  % to make drawing with TikZ
\usetikzlibrary{arrows,positioning,shapes}
\usepackage{tcolorbox}
\usepackage{color}
\usepackage{latexsym}  
\usepackage{tikz}
\usepackage[ruled,vlined,linesnumbered]{algorithm2e}

\title{Online (Non-)Convex Learning via Tempered Optimism}
\author{\name Maxime Haddouche \email maxime.haddouche@inria.fr \\
      \addr Inria, Ecole Normale Supérieure, PSL Research University, France
      \AND
       \name Olivier Wintenberger \email olivier.wintenberger@sorbonne-universite.fr \\
      \addr Sorbonne Université, France
      \AND
      \name Benjamin Guedj \email benjamin.guedj@inria.fr\\
      \addr Inria and University College London, France and UK
      }

\usepackage[english]{babel} %or french
\RequirePackage[%
  pdfstartview=FitH,%
  breaklinks=true,%
  bookmarks=true,%
  colorlinks=true,%
  linkcolor= blue,
  anchorcolor=blue,%
  citecolor=blue,
  filecolor=blue,%
  menucolor=blue,%
  urlcolor=blue%
  ]{hyperref}

\newcommand{\xbf}{\ensuremath{\mathbf{x}}}

\newcommand{\ybf}{\ensuremath{\mathbf{y}}}

\newcommand{\zbf}{\ensuremath{\mathbf{z}}}

\newcommand{\Ebb}{\ensuremath{\mathbb{E}}}

\newcommand{\RA}{\right\rangle}
\newcommand{\LA}{\left\langle}
\newcommand{\LB}{\left[}
\newcommand{\RB}{\right]}

\newcommand{\LP}{\left(}
\newcommand{\RP}{\right)}

\newcommand{\ie}{\textit{i.e.}\xspace}

\newcommand{\eg}{\textit{e.g.}\xspace}

\newcommand{\wrt}{\textit{w.r.t.}\xspace}

\DeclareMathOperator*{\EE}{\Ebb}

\newcommand{\x}{\xbf}

\newcommand{\y}{\ybf}

\newcommand{\z}{\zbf}

\usepackage{cleveref}
\crefname{assumption}{Assumption}{Assumptions}
\crefname{equation}{Eq.}{Eqs.}
\crefname{table}{Table}{Tables}
\crefname{section}{Sec.}{Secs.}
\crefname{theorem}{Thm.}{Thms.}
\crefname{lemma}{Lemma}{Lemmas}
\crefname{corollary}{Cor.}{Cors.}
\crefname{example}{Example}{Examples}
\crefname{exo}{Exercise}{Exercises}
\crefname{appendix}{Appendix}{Appendices}
\crefname{remark}{Remark}{Remark}
\crefname{figure}{Fig.}{Figs.}

\newtheorem{theorem}{Theorem}[section]
\newtheorem{definition}[theorem]{Definition}

\newtheorem{lemma}[theorem]{Lemma}
\newtheorem{proposition}[theorem]{Proposition}

\newtheorem{remark}[theorem]{Remark}

\usepackage{dsfont}
\usepackage{wrapfig}

\begin{document}

\maketitle

\begin{abstract}
  Optimistic Online Learning aims to exploit experts conveying reliable information to predict the future. However, such implicit optimism may be challenged when it comes to practical crafting of such experts. A fundamental example consists in approximating a minimiser of the current problem and use it as expert. In the context of dynamic environments, such an expert only conveys partially relevant information as it may lead to overfitting. To tackle this issue, we introduce in this work the \emph{optimistically tempered} (OT) online learning framework designed to handle such imperfect experts. As a first contribution, we show that tempered optimism is a fruitful paradigm for Online Non-Convex Learning by proposing simple, yet powerful modification of Online Gradient and Mirror Descent. Second, we derive a second OT algorithm for convex losses and third, evaluate the practical efficiency of tempered optimism on real-life datasets and a toy experiment. 
  \end{abstract}

\section{Introduction}

Online learning (OL) is a paradigm in which data is processed sequentially, either because the practitionner does not collect all data prior to analysis or because the dataset dynamically evolves through time, or simply because handling batch of data is numerically too demanding. From the seminal work of \citet{zinkevich2003online}, which proposed an online version of the celebrated gradient descent algorithm, OL has been at the core of many contributions (we refer to \citealp{hazan2007logarithmic, duchi2011adaptive, rakhlin2013predict} for an overview).
The classical performance criterion of an online learning algorithm is the \emph{static regret}. For a sequence of loss functions $(\ell_t : \mathcal{K}\rightarrow \mathbb{R})_{t\geq 1}$ sequentially given by Nature ($\mathcal{K}$ is the predictor space), the static regret compares the efficiency of a predictors sequence $\hat{\xbf}=(\hat{\xbf}_t)_{t\geq 1}$ to the best fixed strategy:
$\operatorname{S-Regret}_T(\hat{\xbf}) = \sum_{t=1}^T \ell_t(\hat{\xbf}_t) - \inf_{\xbf \in \mathcal{K}} \sum_{t=1}^T \ell_t(\xbf),$ $T>0.$
Classical upper bounds on static regret involve a sub-linear rate. For instance, \citet{zinkevich2003online} proposed a $\mathcal{O}(\sqrt{T})$ bound for Online Gradient Descent (OGD) which is valid for convex losses. \citet{hazan2007logarithmic} proved a $\mathcal{O}(d\log(T))$ rate for the Online Newton Step (ONS) algorithm with exp-concave losses.

\textbf{Dynamic Regret.}
Static regret may not be sufficient to assert the efficiency of an online algorithm when evolving in dynamic environments. Hence, the notion of \emph{dynamic regret} introduced by  \citet{zinkevich2003online} and further developed by \citet{hall2013dynamical}, among others. For any sequence $\hat{\xbf}=(\hat{\xbf}_t)_{t\geq 1}$, the worst-case dynamic regret is given by
\[ \operatorname{D-Regret}_T(\hat{\xbf})=  \sum_{t=1}^T \ell_t(\hat{\xbf}_t) -  \sum_{t=1}^T\inf_{\xbf \in \mathcal{K}} \ell_t(\xbf)\,,\qquad T\ge 1\,.  \]
Dynamic regret has attracted many studies recently 
 \citep{besbes2015non,jadbabaie2015online,yang2016tracking,zhang2017improved,zhang2018strong,zhao2021improved}
 but also for any comparator sequence (universal dynamic regret, as in \citealp{zhao2020dyn}).
Those works have established various upper bounds which depend on measures of the cumulative distance between successive optima. For any horizon $T\ge 1$, for any sequence $\xbf=(\xbf_t)_{t\geq 1}$, \citet{zinkevich2003online} introduced the \emph{path length} to measure this discrepancy $P_T(\xbf)= \sum_{t=1}^{T-1}\| \xbf_{t+1}-\xbf_t \|.$
\citet{zhang2017improved} introduced the \emph{squared path length}: $S_T(\xbf)= \sum_{t=1}^{T-1}\| \xbf_{t+1}-\xbf_t \|^2.$
Finally, the \emph{function variation} has been introduced by \citet{besbes2015non}: for any sequence of losses $(\ell_t)_{t\geq 1}$ (these are provided by the environment),$ V_T^\ell(\xbf)= \sum_{t=1}^{T-1} \sup_{\xbf\in\mathcal{K}}\| \ell_{t+1}(\xbf)- \ell_t(\xbf) \|.$
When using the path length\footnote{similar definitions hold for the squared path length and the function variation.} $P_T^*:= P_T(\xbf^*)$ of the minimisers $\xbf^*=(\xbf_t^*)_{t\geq 1}$, dynamic regret of OGD
is at most $\mathcal{O}(\sqrt{T(1+P_T^*)})$ for convex functions \citep{zinkevich2003online,yang2016tracking}. We similarly define $S_T^*:= S_T(\xbf^*)$.
\\
For strongly convex and smooth functions, \citet{mokhtari2016online} established that the dynamic regret is $\mathcal{O}(P_T^*)$. \citet{zhang2017improved} introduced the Online Multiple Gradient Descent (OMGD) and the Online Multiple Newton Update (OMNU) which achieved a $\mathcal{O}(\min(P_T^*,S_T^*))$ dynamic regret. \citet{yang2016tracking} showed that the $\mathcal{O}(P_T^*)$ rate is
also reached for convex and smooth functions under the assumption that all minimisers lie onto the interior of a convex set of interest. \citet{besbes2015non} proved a $\mathcal{O}(T^{2/3}(V_T^*)^{1/3})$ dynamic regret for OGD with a restarting strategy.
Finally, \citet{baby2019online} improved the rate to $\mathcal{O}(T^{1/3}(V_T^*)^{2/3})$ for 1-dimensional square loss with filtering techniques. Note that all the aforementioned results assume implicitly access to $P_T^*$, $S_T^*$, $V_T^*$ and that a notion of \emph{universal dynamic regret} has been studied by  \citet{zhang2018adaptive, zhao2020dyn,zhao2022non} to compete with any $P_T(\xbf)$, $S_T(\xbf)$, $V_T(\xbf)$ rather than $P_T^*$, $S_T^*$, $V_T^*$.
\\
\textbf{Optimistic online learning.}
Optimistic online learning exploits, at each time step, a (possibly) history-dependent additional information provided by an expert. Being optimistic in this context is assuming implicitly that the experts are reliable and can be exploited within an optimisation procedure. Optimistic online learning can be traced back to \citet{hazan2010extracting,chiang2012online} and has been further developed by \citet{rakhlin2013predict,rakhlin2013games} which introduced the celebrated Optimistic Mirror Descent (OptMD).
Those works involved static regret bound highlighting the benefits of experts. \citet{jadbabaie2015online} bridged the gap between dynamic regret and optimistic online learning by providing an adaptive version of OptMD allowing dynamic regret bounds for bounded convex functions. The optimism paradigm has been later developed in \citet{mohri2016accelerating,joulani2020modular,chen2024optimistic}, among others.

\textbf{How to obtain such experts?} While optimistic algorithms come with strong regret guarantees, the question of crafting such desirable experts remains open. In \cite{rakhlin2013games,rakhlin2013predict}, authors assume optimistically to have directly access to an expert in the gradient space, which is also the implicit prerequisite in \cite{jadbabaie2015online}. Unfortunately, there is no clear way to obtain such experts. Alternatively, a natural method to get some additional knowledge would be, at time $t$, to perform an auxiliary offline procedure to catch a (local) minimiser of $\ell_t$ and consider as an expert this minimiser, which then lies in the predictor space. While more practical, this notion of expert differs fundamentally from the one on the gradient space: local minimisers have null gradients and then give no useful information in the gradient space. While not stated as an optimistic algorithm, this idea already appeared in the OMGD algorithm of \cite{zhang2017improved}. However, this procedure is naive from an optimistic perspective as it fully relies on the expert which can overfit on the current problem and then predict poorly in the context of dynamic environments. 

\textbf{Optimistically Tempered Online Learning.}
Then, optimism seems to be subject to a dilemma: we either assume to have experts on the gradient space without an explicit way of crafting them, or naively rely on explicit experts overfitting on past data. To solve this apparent problem, we formalise a novel learning paradigm, namely \emph{Optimistically Tempered Online Learning (OT-OL)}, which is about dealing with experts providing only partially useful information. This is the case, \eg, when the expert at time $t$ is a local minimiser of $\ell_t$: it conveys some information if $\ell_{t+1}$ is correlated to $\ell_{t}$  but may also lead to overfitting.   
\\
In other words, OT-OL weakens the confidence assumption on the experts than Optimistic OL. While not denoted as Tempered Optimism, some OT-OL algorithms already emerged for linear losses in \citet{bhaskara2020online} as a follow-up of \citet{dekel2017online}. It is also possible to interpret online model selection (\emph{e.g.} \citealp{orabona2014simul,wintenberger2017boa}) as OT-OL algorithms as it attenuates the confidence  in a single expert relatively to others. While those results are limited to linear or convex losses, we aim, in this work, to derive OT algorithms, valid for possibly non-convex losses.

\textbf{Contributions and outline.} We investigate Tempered Optimism through original learning algorithms and associated dynamic regret bounds. In \Cref{sec: nonconvex} we show unexpectedly that Tempered Optimism is a fruitful paradigm to maintain sublinear dynamic regret bounds for non-convex losses yielding the following conclusion: \emph{Tempered Optimism compensates the absence of convexity}. To prove so, we unveil \emph{Optimistically-Tempered Online Gradient Descent (OT-OGD)}, based on simple, yet powerful modification of OGD to reach dimension-free dynamic regret bounds. Using the same techniques, we also recover an optimistically tempered version of Online Mirror Descent. In \Cref{sec: convex}, we focus on the convex case to craft explicitly our experts. To exploit them, we design namely \emph{Dynamic OGD (D-OGD)}, exploiting differently tempered optimism and also enjoying dynamic regret bounds. Our algorithm is flexible enough to also enjoy robustness guarantees \wrt to the randomness of the environment.  
Finally, we perform experiments in \Cref{sec: experiments} to assess the practical relevance of Tempered Optimism. Precisely, we test D-OGD on several real-life datasets and a toy experiment illustrating the benefits of D-OGD compared to existing methods.

\section{Online Non-Convex Learning via Tempered Optimism}
\label{sec: nonconvex}

\paragraph{Framework.}
In this work (unless explicitly precised), we use the following mathematical objects and their associated assumptions. First, the set of predictors $\mathcal{K}\subseteq \mathbb{R}^d$ is a closed convex set with finite diameter $D$. We denote by $\Pi_{\mathcal{K}}$ the orthogonal projection on $\mathcal{K}$. Second, we denote by $||.||$ the Euclidean norm on $\mathbb{R}^d$ and assume all gradients are bounded by some constant $G$ \wrt this norm:  $\forall t\geq1, \xbf\in\mathcal{K},  ||\nabla\ell_t(\xbf)||\leq G$. Finally, for an horizon $T>0$, denote by $\LP \mathcal{F}_t\RP_{t\in 1\cdots T}$ a filtration adapted to $(\ell_t)_{t\in 1\cdots T}$. For each $t$, let $\nu_{t}\in\mathcal{K}$ a $\mathcal{F}_t$-measurable expert, available to solve $\ell_{t+1}$. 

While optimistic online learning mostly stands for linear or convex losses \citep{rakhlin2013predict,rakhlin2013games,jadbabaie2015online,flaspohler2021online}, it has been recently shown that optimism is compatible with non-convex losses \citep{suggala2020online}. However, optimism is involved in this case as a complementary result, enriching an independent learning algorithm. In this section, we tackle the following question: can we use experts to get competitive guarantees for non-convex learning problems? 
\\
This is a challenging question as optimistic algorithms, \eg OptMD \citep{rakhlin2013games}, plug the expert as a supplementary gradient/mirror step: a mechanism whose theoretical guarantees are intrinsically linked to the convex nature of losses. We provide a positive answer: \emph{Tempered Optimism compensates lack of convexity} by a careful interpolation between the benefits of the gradient step and those of the expert. To do so, we first need to identify desirable properties satisfied by the experts.

\textbf{Identifying the experts.} As precised earlier, we focus at time $t$ on obtaining $\nu_{t}$ nearly minimising $\ell_{t+1}$. To do so, we assume to have access to an \emph{approximate optimisation oracle}, a notion which emerged in \citet{agarwal2019learning} and later expanded in \citet{suggala2020online,xu2024online}. 
\begin{definition}
  A $\xi$-approximate optimisation oracle $\mathbf{O}$  takes as input any function $\ell$ and returns an approximate minimizer $ \mathbf{O}(\ell)\in\mathcal{K}$ such that
 \begin{equation}
     \label{eq:propertyoracle}
     \ell\left(\mathbf{O}(\ell)\right) \leq \inf_{\xbf\in\mathcal{K}} \ell(\mu)+\xi.
 \end{equation}
\end{definition}

The notion of approximate oracle in non-convex learning translates that, contrary to the convex case, we cannot have subroutines provably reaching a global minimiser of $\ell_t$ at time $t$ (\eg offline gradient descent) and that we can only hope to reach local minimiser whose quality is determined by $\xi$. However, this assumption remains credible: indeed, consider $\mathbf{O}$ to be the SGD algorithm with $K$ iterations (\ie at each time steps we perform $K$ SGD steps to get $\nu_{t}$), then, even when deep nets are involved, it is plausible that $\mathbf{O}$ will reach a local minimiser. Indeed, various works studied SGD for nonconvex losses and identified regimes that prove either on average \citep{ghadimi2013stochastic}, or almost sure convergence \citep{mertikopoulos2020almost,patel2021stochastic,cutkosky2023optimal} to a stationary point. Such points can be local/global minima or saddle points. More recent works showed that there were learning scenario were saddle points were avoided. For instance in \citep{mertikopoulos2020almost}, it was shown that the trajectories of SGD avoid all strict saddle manifolds – i.\eg, sets of critical points x with at least one negative Hessian eigenvalue. Such
manifolds include connected sets of non-isolated saddle points that are
common in the loss landscapes of overparametrized neural networks \citep{li2018visual}. Beyond SGD, almost surely convergence of the stochastic Riemannian Robbins–Monro method (a general template including various algorithms) to a local or global minimum has also been proven in \citep{hsieh2023riemannian}.
We now can state our novel algorithms

\paragraph{Optimistically Tempered Online Gradient Descent.}

Our first algorithm lies in \Cref{alg: OT_OGD}.

\begin{algorithm}[!h]
 \SetAlgoLined
 \SetKwInOut{Initialisation}{Initialisation}
 \SetKwInOut{Parameter}{Parameters}
 \Parameter{Horizon $T$, step-size $\eta$, approximate optimisation oracle $\mathbf{O}$ }
 \Initialisation{Initial point $\xbf_1\in\mathcal{K}$}
\textbf{For} $t$ in $\{1,\dots, T\}$:\\
\hspace{5mm}Observe $\ell_t$. Get $\nu_{t}=\mathbf{O}(\ell_t)$ \\
\hspace{5mm}Draw independently $u_{t} \sim \text{Unif}([0,1])$  \\
\hspace{5mm}Update $\hat{\xbf}_{t+1} = \Pi_{\mathcal{K}}\LP\hat{\xbf}_{t} - \eta \nabla \ell_t \left(\nu_{t} + u_t (\hat{\xbf}_{t} - \nu_{t})\right)\RP.$ \\
 \caption{Optimistically Tempered Online Gradient Descent (OT-OGD)}
 \label{alg: OT_OGD}
 \end{algorithm}

 In \Cref{alg: OT_OGD}, the gradient term has been defined by a random interpolation between $\nu_{t}$ and $\hat{\xbf}_t$ in line 4. This step fits the optimistically tempered paradigm: we aim to incorporate the benefits of experts without relying too much on it. From a technical perspective, the reason behind this interpolation is that it allows involving the fundamental theorem of calculus to avoid the convexity property; an idea that has already been developed for online-to-nonconvex conversions \citep{cutkosky2023optimal}. \Cref{alg: OT_OGD} enjoys the following regret bound, holding for nonconvex losses.

\begin{theorem}
\label{th: OT-OGD}
        Let $\LP \hat{\xbf}_{t} \RP_{t\geq 1}$ be the output of \Cref{alg: OT_OGD} with step $\eta>0$ and $\xi$-approximate optimisation oracle $\mathbf{O}$. Then,
    $$
\EE \LB\sum_{t=1}^T \ell_t(\hat{\xbf}_t) - \sum_{t=1}^T \inf_{\xbf\in\mathcal{K}}\ell_t(\xbf)\RB \leq \frac{1}{2\eta}\left(D^2 + 2DP_T(\nu)\right) + \frac{\eta }{2}G^2 T + \xi T,
    $$
    where the expectation is taken over the sequence $(u_1, \ldots, u_T)$ in \Cref{alg: OT_OGD}.
\end{theorem}

 Assuming $\xi= T^{-1/2}$ (\ie, assuming that the approximation gives qualitative local minima, similarly to \citealp{suggala2020online}) and optimising $\eta$ would yield a bound of $\mathcal{O}\LP \sqrt{T(1+P_T(\nu))}\RP$ matching the optimal rate for the convex case \citep{yang2016tracking} at the cost of switching the optimal path $P_T^*$ ill-defined in the nonconvex case, to $P_T(\nu)$, being a path between local minima. Furthermore, we can be slightly more sharp by not bounding uniformly the gradients by $G^2$ (see the proof below). In this case, the optimal rate would be $\mathcal{O}\LP \mathbb{E}\LB \sqrt{\sum_{t=1}^T \left\| \nabla \ell_t (\nu_{t} + u_t (\hat{\xbf}_{t} - \nu_{t})) \right\|^2 (1+P_T(\nu))}\RP \RB$, meaning that vanishing gradients can tighten the bound. This tightened bound highlights that, as long as the sum of gradients vanishes faster than $P_T(\nu)$, OT-OGD enjoys a faster convergence guarantee than the naive strategy $\hat{\x}_{t+1}= \nu_t$, giving full confidence on the expert and enjoying a regret bound of $GP_T(\nu) + \xi T$. Note also that for  extremely adversarial problems ($P_T(\nu)= \mathcal{O}(T^\beta),\beta>1$), \Cref{th: OT-OGD} also outperforms the naive strategy. 

 \textbf{Role of the path length.} Note that in our theorem, we consider the path $P_T(\nu)$ with respect to our experts. This stems from the fact that in nonconvex learning problems, there is no uniqueness of the global minimiser (or maybe no minimiser at all), making $P_T^*$ hard to define. If $\nu_t = \mathbf{O}(\ell_t)$, $P_T(\nu)$ still conveys information on Nature via a path between local minima. Note also that if we assume having an optimisation oracle with $\xi=0$ then $P_T(\nu)$ coincides with $P_T^*$.  

 \textbf{Comparison with literature} While \Cref{th: OT-OGD} is not the first regret bound for nonconvex losses, it is to our knowledge, the first dynamic regret bound for nonconvex losses avoiding an explicit dependency in the dimension $d$ of the predictor space. Indeed, \citet[Corollary 3]{krichene2015hedge} managed to get a rate of $\mathcal{O}(\sqrt{d T})$ while \citet[Theorem 1]{agarwal2019learning} involved a polynomial dependency in $d$. This dependency has been later made explicit in \citet[Theorem 1]{suggala2020online}, being $\mathcal{O}\LP d^{3/2} \sqrt{T}\RP$, even for their optimistic algorithm. Furthermore, those results hold for static regret. More recently, \citet{xu2024online} managed to reach dynamic regret bounds for nonconvex losses with the rate of $\mathcal{O}d\LP T^{2/3} (1+V_T)^{1/3} \RP$, which is different of our $\mathcal{O}\LP T^{1/2} (1+P_T)^{1/2} \RP$ as long as $\xi= \mathcal{O}(T^{-1/2})$. Furthermore, other line of works such as \citet{cutkosky2023optimal} managed to get theoretical guarantees for nonconvex learning problems with no explicit dependency in the dimension. This comes at the cost of considering weaker notion of regrets such as stationary points. This shows the strength of tempered optimism combined with approximate optimisation oracle: we avoid any explicit dependency in $d$ to match the optimal rate for convex functions.

\begin{proof} 
   In what follows, we write for any $t\in\{1\cdots T\}$, $\mathbb{E}_{u_t}$ the expectation under $u_t\sim \text{Unif}([0,1])$ and $\mathbb{E}_{u_{1:T}}$ the expectation under the joint distribution $(u_1, \ldots, u_T)\sim \text{Unif}([0,1])^{\otimes T}$. 
   First, we exploit that $\nu_{t} =\mathbf{0}(\ell_t)$ to get:
   \begin{align*}
    \sum_{t=1}^{T}\ell_t (\hat{\xbf}_{t}) - \inf_{\xbf\in\mathcal{K}}\ell_t(\xbf)  & = \sum_{t=1}^{T}\ell_t (\hat{\xbf}_{t}) - \ell_t (\nu_{t}) + \sum_{t=1}^{T}\ell_t (\nu_{t}) - \inf_{\xbf\in\mathcal{K}}\ell_t(\xbf) \\
    & \leq \sum_{t=1}^{T}\ell_t (\hat{\xbf}_{t}) - \ell_t (\nu_{t}) + \xi T.
   \end{align*}
   
   To deal with the remaining sum, we have, by the fundamental theorem of calculus:
    \begin{align}
        \sum_{t=1}^{T}\ell_t (\hat{\xbf}_{t}) - \ell_t (\nu_{t}) &= \sum_{t=1}^T\int_{u=0}^{1} \LA \nabla\ell_t(\nu_{t} + u(\hat{\xbf}_{t} - \nu_{t})),\hat{\xbf}_{t} - \nu_{t}\RA du \notag \\
        &= \sum_{t=1}^{T}\LA\mathbb{E}_{u_t}\LB\nabla \ell_t (\nu_{t} + u_t (\hat{\xbf}_{t} - \nu_{t}))\RB, \hat{\xbf}_{t} - \nu_{t}\RA \notag\\
        &=\sum_{t=1}^{T}\LA \mathbb{E}_{u_t} \LB\tilde{\nabla}_t (u_t)\RB, \hat{\xbf}_{t} - \nu_{t}\RA ,\label{eq:expectedgradient}
    \end{align}
    where $\tilde{\nabla}(u_t) :=\nabla \ell_t (\nu_{t} + u_t (\hat{\xbf}_{t} - \nu_{t}))$. To control \eqref{eq:expectedgradient} we note that 
    \begin{align}
    \mathbb{E}_{u_t} \LB\|\hat{\xbf}_{t+1}-\nu_{t}\|^2\RB&=\mathbb{E}_{u_t}\|{\Pi_{\mathcal{K}}[\hat{\xbf}_{t}  -\eta \tilde{\nabla}_t (u_t)]-\nu_{t}\|^2 } \notag \\
    &\leq \mathbb{E}_{u_t}\LB\|\hat{\xbf}_{t} - \nu_{t} - \eta\tilde{\nabla}_t (u_t)\|^2\RB \notag\\
    &\leq \|\hat{\xbf}_{t} - \nu_{t}\|^2 - 2\eta\LA\mathbb{E}_{u_t}\LB\tilde{\nabla}_t (u_t)\RB,\hat{\xbf}_{t} - \nu_{t} \RA + \eta^2 G^2 \label{eq:decompositionsquare} \\
    \intertext{where we used $\|\tilde{\nabla}_t (u_t)\|^2 \leq G^2$. Re-arranging \eqref{eq:decompositionsquare}  and taking the expectation over $(u_1, \ldots, u_T)$ yields:}
\mathbb{E}_{u_{1:T}}\LB\LA\tilde{\nabla}_t (u_t),\hat{\xbf}_{t} - \nu_{t} \RA \RB&\leq \mathbb{E}_{u_{1:T}}\LB \frac{1}{2\eta}\left(\|\hat{\xbf}_{t} - \nu_{t}\|^2 -\|\hat{\xbf}_{t+1}-\nu_{t}\|^2 \right)+ \frac{\eta G^2}{2} \RB,\label{eq:boundinstantaneous}
    \end{align}

    Now notice that, for any $t\in\{1\cdots m-1\}$, 
    \begin{align} 
\|\hat{\xbf}_{t+1}-\nu_{t+1}\|^2 &= \|\hat{\xbf}_{t+1}-\nu_{t}\|^2 + \|\hat{\xbf}_{t+1}-\nu_{t+1}\|^2-\|\hat{\xbf}_{t+1}-\nu_{t}\|^2 \notag \\
&=  \|\hat{\xbf}_{t+1}-\nu_{t}\|^2 \notag\\
&+ (\|\hat{\xbf}_{t+1}-\nu_{t+1}\| + \|\hat{\xbf}_{t+1}-\nu_{t}\|)(\|\hat{\xbf}_{t+1}-\nu_{t+1}\| - \|\hat{\xbf}_{t+1}-\nu_{t}\|) \notag\\
&\leq  \|\hat{\xbf}_{t+1}-\nu_{t}\|^2 + 2D\|\nu_{t+1}-\nu_{t}\| \notag,
\intertext{where we have used that $\mathcal{K}$ has a finite diameter $D$ and the reversed triangular inequality in the last line. It follows that:}
-\|\hat{\xbf}_{t+1}-\nu_{t}\|^2 & \leq -\|\hat{\xbf}_{t+1}-\nu_{t+1}\|^2 + 2D\|\nu_{t+1}-\nu_{t}\|
\label{eq:controltplusone}\
    \end{align}
    Plugging \eqref{eq:controltplusone} into \eqref{eq:boundinstantaneous} then gives: 
    \begin{align*}
        \mathbb{E}_{u_{1:T}}\LB\LA\tilde{\nabla}_t (u_t),\hat{\xbf}_{t} - \nu_{t}\RA\RB&\leq \mathbb{E}_{u_{1:T}}\left[\frac{1}{2\eta}\left(\|\hat{\xbf}_{t} - \nu_{t}\|^2 -\|\hat{\xbf}_{t+1}-\nu_{t+1}\|^2 \right) +\frac{1}{\eta}D\|\nu_{t+1}-\nu_{t}\|+ \frac{\eta G^2}{2}\right].
    \end{align*}
    
    Thus, summing over $T$ and plugging the resulting sum in \eqref{eq:expectedgradient} finally gives:
    \begin{align*}
        \mathbb{E}_{u_{1:T}}\LB\sum_{t=1}^{T}\ell_t (\hat{\xbf}_{t}) - \ell_t (\nu_{t})\RB & \leq  \mathbb{E}_{u_{1:T}}\left[\frac{\|h_1 - \nu_{0}\|^2}{2\eta} +\sum_{t=1}^T\frac{1}{\eta}D\|\nu_{t+1}-\nu_{t}\|+ \sum_{t=1}^T\frac{\eta G^2}{2}\right]. \\
    & \leq  \frac{D^2}{2\eta}+\frac{D}{\eta}\sum_{t=1}^{T}\|\nu_{t+1}-\nu_{t}\|+ \frac{\eta  G^2 T}{2}. 
    \end{align*}
\end{proof}

\paragraph{A theoretical extension: Optimistically Tempered Online Mirror Descent.}
To reach regret bounds for OT-OGD, our main theoretical trick has been to exploit the fundamental theorem of calculus to involve gradient terms without using convexity. We show below that this trick goes beyond the particular case of OGD and present \emph{Optimistically Tempered Online Mirror Descent (OT-OMD)} which acts as a generalisation of OT-OGD, still valid for nonconvex losses.

We consider regularisation functions, denoted $R: \mathcal{K} \mapsto \mathbb{R}$, which are $\alpha$-strongly convex with finite diameter $D_R^2 := \operatorname{argmax}_{\x,\y} R(\x)-R(\y) <+\infty$.
We also assume that the regularisation functions are twice differentiable over $\mathcal{K}$.
We then define the \emph{Bregman divergence} with regularization $R$ as
$B_R(\xbf \mid \ybf): =R(\xbf)-R(\ybf)-\nabla R(\ybf)^{\top}(\xbf-\ybf) .$

\begin{algorithm}[!h]
 \SetAlgoLined
 \SetKwInOut{Initialisation}{Initialisation}
 \SetKwInOut{Parameter}{Parameters}
 \Parameter{Horizon $T$, step-size $\eta$, approximate optimisation oracle $\mathbf{O}$, regulariser $R$ }
 \Initialisation{Initial point $\hat{\xbf}_1 = \nu_{0}\in \mathcal{K}$.}
\textbf{For} $t$ in $\{1,\dots, T\}$:\\
\hspace{5mm} Observe $\ell_t$. Get $\nu_{t}=\mathbf{O}(\ell_t)$ \\
\hspace{5mm} Draw independently $u_t\sim \text{Unif}([0,1])$ and compute $\nabla_t := \nabla \ell_t\LP u_t \hat{\xbf}_t + (1-u_t)\nu_{t} \RP.$ \\
\hspace{5mm} Update $ \y_{t+1} = \operatorname{argmin}_{\y} \eta\langle \nabla_t, \y\rangle + B_R(\y,\hat{\xbf}_t) $ \\
\hspace{5mm} Projection step: $\hat{\xbf}_{t+1}=\operatorname{argmin}_{\x\in \mathcal{K}} B_R(\x,\y_{t+1})$.
 \caption{Optimistically Tempered Online Mirror Descent (OT-OMD).}
 \label{alg: OT-OMD}
 \end{algorithm}

 We have the following regret bound: 

 \begin{theorem}
  \label{th: OT-OMD}
  Let $\mathbf{O}$ be a $\xi$-approximate optimisation oracle assume that there exists a bound $G_R$ on the local norms defined in \Cref{sec: technical_b}. Let $(\hat{\xbf}_t)_{t\geq1}$ be the output of OT-OMD (\Cref{alg: OT-OMD}) with step-size $\eta$, then : 

  \begin{align*}
    \mathbb{E}_{u_1,\cdots,u_T}\LB \sum_{t=1}^{T}  \ell_t(\hat{\xbf}_t) - \sum_{t=1}^T \inf_{\xbf\in\mathcal{K}}\ell_t(\xbf)\RB & \leq \frac{ D_R^2 + 2G^*P_T(\nu)}{2\eta} + \frac{\eta}{2}G_R^2T  + \xi T,
  \end{align*}
  where $G^*= \max_{t\in\{1\cdots T\}} \mathbb{E}_{u_1\cdots u_T} \LB\|\nabla R(\hat{\x}_t)\| \RB$.

 \end{theorem} 

Proof is deferred to \Cref{sec: proof-nonconvex}.  Similarly to OT-OGD, we can obtain a slighly sharper version if we do not crudely upper bound the gradients by $G_R^2$. In this case, the factor $G^2T$ would tighten into $\sum_{t=1}^{T}\mathbb{E}_{u_1,\cdots,u_T}\LB (\|\nabla_t\|_t^*)^2 \RB$, where $\|.\|_t^*$ denotes a dual norm at time $t$, see \Cref{sec: technical_b}. Similarly to OT-OGD, OT-OMD enjoys a dimension-free regret bound which extend the original static regret bound for linear losses and Optimistic OMD \citep{rakhlin2013predict}. 

 \paragraph{Application: Tracking the best expert in nonconvex problems.} We consider the setting where $K$ experts $(\theta_t^1,\cdots,\theta_t^K)\in \mathcal{K}^K$ are available at each time $t$ and we aim to find the best aggregation of experts $\hat{\x}_t\in \mathcal{S}^K$ (the simplex in $\mathbb{R}^K$) to answer a non-convex problem $\ell_t$. In this case, we can recover the Hedge algorithm \citep{hazan2019introduction} by considering $R(\x) = \x \log(\x) = \sum_{i=1}^K \x_i\log(\x_i), \x\in \mathcal{S}^K$, where the logarithm is taken element-wise. $R$ is a $1$-strongly convex function with respect to the $1$-norm. Then, assuming that the costs per individual expert are in the range $[0,1]$. OT-OMD can be applied with $D_R^2=\log(K),G_R=1$ (see page 82 of \citealp{hazan2019introduction}). In this case, considering at each time step $\nu_{t}\in \operatorname{argmin}_{\theta\in (\theta_t^1,\cdots,\theta_t^K)} \ell_t(\theta)$ gives the following regret bound. 
\begin{align}
  \label{eq: expert_tracking}
    \mathbb{E}_{u_1,\cdots,u_T}\LB \sum_{t=1}^{T}  \ell_t(\hat{\xbf}_t) - \sum_{t=1}^T \inf_{j\in \{1\cdots K\}}\ell_t(\theta_t^j)\RB & \leq \frac{ \log(K) + 2G^*P_T(\nu)}{2\eta} + \frac{\eta}{2}T ,
  \end{align}
  where $G^*= \max_{t\in\{1\cdots T\}} \mathbb{E}_{u_1\cdots u_T}\LB\|\nabla R(\hat{\x}_t)\|\RB$. Here, we do not pay the $\xi T$ factor as we directly crafted our expert instead of relying on a offline optimisation oracle. In the non-convex case, a continuised version of the Hedge algorithm have static regret guarantee of $\mathcal{O}(\sqrt{KT\log(T)})$ \citep{krichene2015hedge} against the best static expert aggregation. However, this general result is not fitted for the problem of tracking the best experts which has been largely studied since the seminal work of \citet{herbster2001tracking} for convex or linear losses and often enjoys mild dependencies of $\log(K)$ in the dimension.  Note that in the linear case, searching for the best experts aggregation on the simplex yields a solution on one of the simplex vertices, a finite and identifiable set of predictors in line with what appears in the left-hand side of \Cref{eq: expert_tracking}. For this problem, \citet{cesa-bianchi2012mirror} managed to reach a dynamic regret bound involving a dependency of $\log(K)$ as well as a path $m$ depending on the $\ell_1$ norm. Here, our dynamic regret bound maintains similar nice properties for non-convex losses, with a path depending on the euclidean norm, which is more robust to the dimension. Note however that our path $P_T$ is pondered by a problem-dependent quantity $G^*$, possibly huge and that we only compare ourselves to the best expert at each time step, instead of the best aggregation: two reasonable side effects to obtain expert tracking guarantees for non-convex scenarios.

\section{Tempered Optimism for Online Convex Learning}
\label{sec: convex}

We assume in this section that our loss functions $(\ell_t)_{t\geq 1}$ are convex.
    % \[ \forall (t, \xbf,\xbf_0 )\in \mathbb{N}/\{0\}\times \mathcal{K}^2, \ell_t(\xbf) - \ell_t(\xbf_0) \leq \langle \nabla \ell_t(\xbf), \xbf- \xbf_0 \rangle - \lambda \|\xbf-\xbf_0\|^2.   \]

In \Cref{sec: nonconvex}, a limitation to apply Tempered Optimism was to rely on an approximate optimisation oracle. While this assumption is plausible, it is relevant from a practical perspective to have an explicit control on the quality of our experts. In this section, we exploit the properties of convex losses to explicitly craft those experts and derive a novel Optimistically Tempered algorithm designed exploiting them.

\textbf{Constructing experts.} We explicitly craft our experts as follows: at time $t$, starting from $\hat{\xbf}_t$, we apply $K$ gradient descent steps on $\ell_t$. 
We name this procedure \textsc{Construct} and detail it in \Cref{alg: additional_knowledge_OMGD} of \Cref{sec: construct}. It consists in applying, at time $t$, $K_t>0$ steps of the classical gradient descent algorithm to obtain a good approximation of the last observed minimum. This can be interpreted as an explicit approximate optimisation oracle whose parameter $\xi_t$ can be controlled explicitly via convex optimisation guarantees (\Cref{l: GD_add_know}). Those explicit control and crafting are important for practical implementation.

\subsection{The \textsc{Adjust} algorithm}
We introduce \textsc{Adjust} (\Cref{alg: adjust}) an optimistically tempered subroutine which adjusts a candidate predictor with respect to experts available at time $t$. We first introduce the notion of \emph{performance}.

\begin{definition}
  \label{def: perf}
        We use the notation $\langle x,y\rangle_H:= x^T H y$ to denote the inner product associated to a positive definite matrix $H$.
        For a sequence of experts $\mathbf{\nu}=(\nu_{t})_{t\geq 0}$, a sequence
        $\hat{\xbf}_{temp}= (\hat{\xbf}_{temp,t})_{t\geq 1}\in\mathcal{K}^{\mathbb{N}}$ (the output of a classical online procedure) and for any positive definite matrix $H$, one defines the \emph{performance} at time $t$ of  $\hat{\xbf}_{temp}$ with regards to $\mathbf{\nu},H$ as follows: we set $m_t:= \frac{\nu_{t}+\nu_{t-1}}{2}$ and
        \[ \mathrm{Perf}(t,H,\hat{\xbf}_{temp},\mathbf{\nu}):=  \left\langle \hat{\xbf}_{temp,t+1}- m_t, \nu_{t}-\nu_{t-1}\right\rangle_H. \]
\end{definition}

  For more details about this notion of performance, we refer to \cref{sec: inspiration_perf}.
  
\begin{figure}[!ht]
   \centering
   
    \begin{subfigure}[b]{0.295\textwidth}
     \centering
     \includegraphics[width=\textwidth]{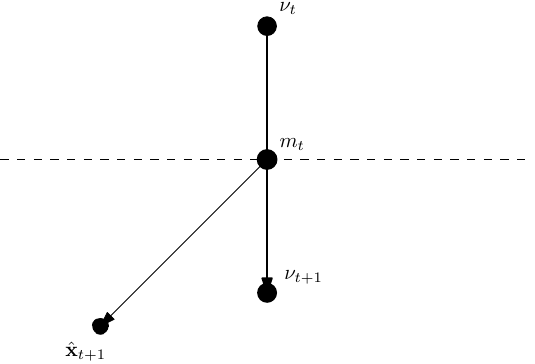}
  \end{subfigure}~
  \begin{subfigure}[b]{0.3\textwidth}
    \centering
    \includegraphics[width=\textwidth]{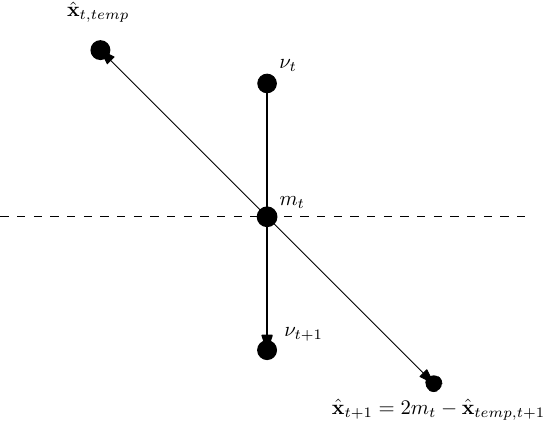}
 \end{subfigure}
   \caption{Output of \textsc{Adjust} when: {\it (left)} performance is positve, {\it (right)} performance is negative.}
   \label{fig: adjust explanation}
 \end{figure}

  \textbf{Understanding the performance.} At time $t$, the performance exploits the expert $\nu$ through $m_t$ and $\nu_{t}-\nu_{t-1}$. The first term is new to the best of our knowledge while the second acts as a proxy to recover an information on the gradient space, as in most of the Optimistic OL literature \citep{rakhlin2013predict}. 
This point is also highlighted in \cite{jadbabaie2015online} as their path $D_T$ focuses on the distance between experts and the gradients of their predictors.
\\
We now state the algorithm \textsc{Adjust} (\Cref{alg: adjust}) which takes as input $\hat{\xbf}_{temp}$, $\nu$, $H,t$ as defined in \Cref{def: perf} and outputs an updated predictor $\hat{\xbf}_{t+1}$. We denote by $\Pi_{\mathcal{K,H}}$ the projection over the closed convex set $\mathcal{K}$ with respect to the distance induced by $\langle.,.\rangle_H$. We illustrate in \cref{fig: adjust explanation} what \textsc{Adjust} concretely performs when $H=\mathbf{I}_2$ and $\mathcal{K}=\mathbb{R}^2$.

\begin{algorithm}[!h]
 \SetAlgoLined
 \SetKwInOut{Initialisation}{Initialisation}
 \SetKwInOut{Parameter}{Parameters}
 \Parameter{Time $t$, positive definite $H$,
 experts $(\nu_i)_{i=1..t+1}$, candidate $\hat{\xbf}_{temp,t+1}$ }

Set up $m_t = \frac{\nu_{t}+\nu_{t-1}}{2}$. \\
\textbf{If $\mathrm{Perf}(t,H,\hat{\xbf}_{temp},\nu)<0$, then:}\\
    \hspace{5mm}Set $ \hat{\xbf}_{t+1}= \underset{\xbf \in \mathcal{K} }{\arg \min }\left\|2m_t- \hat{\xbf}_{temp,t+1}-\xbf\right\|_{H}^{2} \; := \Pi_{\mathcal{K},H}(2m_t-\hat{\xbf}_{temp,t+1})$\\

    \textbf{Else:}\\
        \hspace{5mm} Set $ \hat{\xbf}_{t+1}= \underset{\xbf \in \mathcal{K} }{\arg \min }\left\|\hat{\xbf}_{temp,t+1}-\xbf\right\|_{H}^{2}  \; := \Pi_{\mathcal{K},H}(\hat{\xbf}_{temp,t+1})$ \\
    \textbf{Return} $\hat{\xbf}_{t+1}$
     \caption{The \textsc{Adjust} algorithm}
     \label{alg: adjust}
     \end{algorithm}

% \begin{wrapfigure}{r}{0.5\textwidth} % r=right, l=left
%     \centering
%     \vspace{-7pt} % shift up if needed
%     \includegraphics[width=0.5\textwidth]{../D-OGD_explanation}
%     \caption{Action of \textsc{Adjust} when performance is negative}
%     \label{fig: adjust explanation}
% \end{wrapfigure}

\Cref{fig: adjust explanation} shows how \Cref{fig: adjust explanation} tempers the impact of the expert: on the left, the performance is positive and only the gradient step is considered. Indeed, involving the experts in this case could lead to overfit as the dynamic provided by $\nu_t-\nu_{t+1}$ points in the same direction as the gradient step. On the right, the performance is negative, \textsc{Adjust} consider that the expert provide original information not contained in the gradient and then adjusts the trajectory of the gradient step. 
\\
The mathematical translation of this analysis is that \textsc{Adjust} makes $\hat{\xbf}_{t+1}$ closer from $\nu_{t}$ than $\nu_{t-1}$, which implies less confidence on the expert than directly involving $\nu_{t}$. This is further developed in \Cref{l: variation_trick}.

\begin{lemma}
  \label{l: variation_trick}
 For all $t\geq 0$, any positive definite $H$, any $\hat{\xbf}_{temp,t+1}$, $\nu_{t}$, $\nu_{t-1}$ defined as in \textsc{Adjust} (\cref{alg: adjust}): we denote by  $\|.\|_H^2$ the norm associated to the scalar product $\langle.,.\rangle_H$. 
 \\ We then have: $\left\|\hat{\xbf}_{t+1}-\nu_{t}\right\|_H^{2} \leq \|\hat{\xbf}_{temp,t+1} - \nu_{t-1}\|_H^2.$
\end{lemma}

Proof is deferred to \Cref{sec: proof_determin_results}.
    In the rest of this section, as we focus on a combination of \textsc{Adjust} and OGD, we only consider the case $H=\mathbf{Id}$ and skip the dependency in $H$ for \textsc{Adjust}. However, note that for more elaborated algorithms, taking an adaptive sequence of $H_t$ at each time steps is crucial to reach Optimistically tempered versions of Online Newton Step \citep{hazan2007logarithmic} and Adagrad \citep{duchi2011adaptive}. We refer to \Cref{sec: additional-OT-algos} for more details.

\subsection{Dynamic Online Gradient Descent}

  We present \emph{Dynamic Online Gradient-Descent} (D-OGD), in \Cref{alg: dynamic_OGD}, which exploits experts in an optimistically tempered manner. Its associated theoretical guarantee for D-Regret is stated in \Cref{th: D_regret_bound_general} and experiments showing its practical efficiency are gathered in \Cref{sec: experiments}.

\begin{algorithm}[!h]
 \SetAlgoLined
 \SetKwInOut{Initialisation}{Initialisation}
 \SetKwInOut{Parameter}{Parameters}
 \Parameter{Horizon $T$, step-sizes $(\eta_t)$ }
 \Initialisation{Initial point $\xbf_1\in\mathcal{K}$ and expert $(\nu_{0})\in\mathcal{K}$ }
\textbf{For} $t$ in $\{1,\dots, T\}$:\\
\hspace{5mm}Update $ \hat{\xbf}_{temp,t+1} = \hat{\xbf}_{t} - \eta_t \nabla \ell_t(\hat{\xbf}_{t}) $ \\
\hspace{5mm}Get $\nu_{t}$, \\
\hspace{5mm}$\hat{\xbf}_{t+1}= \textsc{Adjust}(t,\nu,\hat{\xbf}_{temp,t+1})$ 
 \caption{Projected D-OGD onto a closed convex space $\mathcal{K}$.}
 \label{alg: dynamic_OGD}
 \end{algorithm}

 \Cref{alg: dynamic_OGD} consists in applying a gradient descent step, then observe the expert and (possibly) involve it in the learning procedure through \textsc{Adjust}.   

\begin{theorem}
    \label{th: D_regret_bound_general}
    Assume our predictors $\hat{\xbf}$ are obtained via \Cref{alg: dynamic_OGD} with steps $\mathbf{\eta}=  (\frac{D}{G \sqrt{t}})_{t=1\cdots T}$.
    Also assume experts are crafted via \textsc{Construct} (\Cref{alg: additional_knowledge_OMGD}) with steps $\eta'=(\frac{1}{\lambda j})_{j=1\cdots K_t}$ and $K_t=  t $.
    Then, dynamic regret of D-OGD satisfy :
    \begin{align*}
       \sum_{t=1}^T \ell_t(\hat{\xbf}_t) - \sum_{t=1}^T \inf_{\xbf\in\mathcal{K}}\ell_t(\xbf)
       & \leq  GP_T(\nu)  + \frac{9}{2}GD \sqrt{T}.
    \end{align*}

    Furthermore, this result remains for any expert sequence $\nu$ such that for any $t$, $\ell_t(\nu_{t}) - \inf_{\xbf\in\mathcal{K}}\ell_t(\xbf) = \mathcal{O}(1/\sqrt{t})$.
\end{theorem}

Proof is deferred to \Cref{sec: proof_determin_results}. \Cref{th: D_regret_bound_general} provides a worst-case guarantee for the dynamic regret of D-OGD. An interesting point is that our bound decoupled the influence of the path lengths from the horizon $T$, which is not usual in the literature (\citealp{zinkevich2003online} proposed a bound of $\mathcal{O}(\sqrt{T}(1+P_T))$ later improved in \citet{zhang2018adaptive} in a $\mathcal{O}(\sqrt{T(1+P_T)})$.
\\
\textbf{Time complexity.} \Cref{alg: dynamic_OGD} can be thought independently of \textsc{Construct} when experts are given in advance and satisfy the condition $\ell_t(\nu_{t}) - \inf_{\xbf\in\mathcal{K}}\ell_t(\xbf) = \mathcal{O}(1/\sqrt{t})$. In this case, D-OGD has a $\mathcal{O}(T)$ complexity. The use of \textsc{Construct} within \Cref{alg: dynamic_OGD} comes at the cost of an additional time complexity determined by the number of iterations $K_t$ of \Cref{alg: additional_knowledge_OMGD} at each time step.
In \Cref{th: D_regret_bound_general}, we pay an overall complexity of $\mathcal{O}(T^2)$. Note however that if losses are strongly convex, this time complexity can be improved to $\mathcal{O}(T^{3/2})$ ($K_t = \lceil \sqrt{t} \rceil$, see \Cref{l: GD_add_know}), which is similar to the time complexity of the OMGD algorithm of \cite{zhang2017improved} with step-size $\eta=1/\sqrt{T}$.
\\
\textbf{Comparison with literature.} If $\xbf_t^* \in \operatorname{argmin}_{\xbf\in\mathcal{K}} \ell_t(\xbf)$ is revealed to the learner at time $t+1$, then taking $\nu_{t}=\xbf_t^*$ yields $P_T(\nu)= P_{T-1}(\xbf^*) + ||\xbf_1^* - \nu_{0}||$. This allows us to compare in this case, our results with those of \citet{zhang2017improved}.
Then, our convergence rate is worse than their $\mathcal{O}(\min(P_T(\xbf^*),S_T(\xbf^*)))$ while holding with a single convex assumption (no smoothness or strong convexity are required).  Our result stand also in line with some rates of \citet{zhao2021improved} which requires convexity to get a rate in $\mathcal{O}(P_T)$, while some of their improved rates requires an additional smoothness assumption.\\
 \textbf{Role of the path length.} As in \Cref{sec: nonconvex}, $P_T(\nu)$ \Cref{th: D_regret_bound_general} involves the path between $\nu_t$ which are in this case, explicitly crafted and are good approximation of global minimisers. Note also that, in the context of online convex learning, a simlilar term appeared in \citet[Lemma 3]{rakhlin2013predict}, involving a sum of the distances in the dual space between experts and Nature, \citet{jadbabaie2015online} involves a similar term in the context of dynamic online learning. \\
%Those terms translate the experts impact on the training as well as the interplay between experts and the environment. In our study, we decoupled the evolution of $P_T$ and its interplay with the environment.  Indeed, in our proofs, we used the following regret decomposition: , if $R= \sum_{t=1}^T \ell_t(\hat{\xbf}_t) - \sum_{t=1}^T \ell_t(\xbf_t^*)$, then 
% $R = \underbrace{\sum_{t=1}^T \ell_t(\hat{\xbf}_t) - \sum_{t=1}^T \ell_t(\nu_{t-1})}_{= (A)}-  \underbrace{\sum_{t=1}^T \ell_t(\nu_{t-1}) - \sum_{t=1}^T \ell_t(\nu_{t})}_{=(B)} + \underbrace{\sum_{t=1}^T \ell_t(\nu_{t}) - \sum_{t=1}^T \ell_t(\xbf_t^*)}_{=(C)} .$ 
% (A) is dealt via ADJUST and we choose to separate the terms (B) and (C). Note however, that if we apply directly convexity and bounded gradients on the sum (B)+ (C), we recover a $\mathcal{O}\left(\|\nu_{t-1}-\xbf^*_t\|\right)$ which captures the environment dynamics. However, assuming directly that $\nu_{t-1}$ is closed from $\xbf_t^*$ is optimistic, we then relaxed this assumption to $\ell_t(\nu_{t}) - \ell_t(\xbf_t^*)$ is small (which is more realistic as $\nu_{t}$ is $\mathcal{F}_t$ measurable). 
\textbf{Theoretical and practical follow-ups.} In \Cref{sec: real-life-exp}, we show the impact of D-OGD on various real-life datasets. We show that the Optimistically Tempered approach always matches the best strategy between no optimism at all (OGD) and full commit on the expert (OMGD). From a theoretical standpoint, we show in \Cref{sec: additional-OT-algos} that the use of \textsc{Adjust} goes beyond OGD. We then provide Optimistically Tempered versions of Online Newton Step \citep{hazan2007logarithmic} and Adagrad \citep{duchi2011adaptive}.

\textbf{Towards robustness guarantees.} At the cost of a slightly stronger assumption on the loss (stochastic-exp concacvity, \Cref{sec: proof_proba_results})Furthermore, D-OGD provably satisfies D-Regret bounds on the loss sequence defined for any $t\geq 1$: $\mathbb{E}_{t-1}[\ell_t]= \mathbb{E}[\ell_t \mid \mathcal{F}_{t-1}]$ with $(\mathcal{F}_t)_{t\geq 1}$ a filtration adapted to the environment and $\mathbb{E}_{t-1}[\ell_t]$.
This ensures that our predictors are robust to the randomness of the environment. Thus, we define the \emph{Dynamic Cumulative Risk} (D-C-Risk) (already introduced in \citealp{wintenberger2021stochastic}) as follows: for any predictable\footnote{in the sense that predictors only depend on the past.} sequences $\hat{\xbf}$ and $\xbf$ of predictors (\emph{\emph{i.e.,}}, $\hat{\xbf}_t$ and ${\xbf}_t$ are $\mathcal{F}_{t-1}$ measurable), we denote $L_t= \mathbb{E}_{t-1}[\ell_t]$, $t\ge1$, and $\operatorname{D-C-Risk}_T(\hat{\xbf},\xbf)=  \sum_{t=1}^T L_t(\hat{\xbf}_t) -  \sum_{t=1}^T L_t(\xbf_t),$ $T\ge 1$. We then have the following result.

\begin{theorem}
  \label{th: av-D-OGD}
    Assume $\ell_t$ are $\alpha$-stochastically exp-concave(\textbf{(H2)} in \Cref{sec: proof_proba_results}) and that our predictors $\hat{\xbf}$ are obtained via \Cref{alg: dynamic_OGD} with steps $\mathbf{\eta}=  (\frac{D}{G \sqrt{t}})_{t=1\cdots T}$.
    Also assume experts are crafted via \textsc{Construct} (\Cref{alg: additional_knowledge_OMGD}) with steps $\eta'=(\frac{1}{\lambda j})_{j=1\cdots K_t}$ and $K_t=  t $.
    Then, dynamic cumulative risk satisfies with probability $1-3\delta$, for any $T \geq 1$, for any sequence $(\xbf_t)_{t=1\cdots T}$ such that $\xbf_t$ is $\mathcal{F}_{t-1}$-measurable:
    \begin{align*}
        \sum_{t=1}^T L_t(\hat{\xbf}_t) - \sum_{t=1}^T L_t(\xbf_t)
        & \leq  GP_T(\nu) + \mathcal{O}(\sqrt{T}).
    \end{align*}
    Furthermore, this result remains for any experts $\nu$ such that for any $t$, $\ell_t(\nu_{t}) - \ell_t(\xbf_t^*) = \mathcal{O}(1/\sqrt{t})$.
\end{theorem}

Proof is deferred to \Cref{sec: proof_proba_results}. \Cref{th: av-D-OGD} hold for any predictable sequence of comparators $\xbf$ which are not involved on the upper bound. This maintains a fully empirical upper bound as predictable sequences are often unknown due to their dependency on the conditional distribution of data.
We maintain the same convergence rate as \Cref{th: D_regret_bound_general} while shifting D-Regret for D-C-Regret, at the cost of a stochastic exp-concavity assumption. 
As long as paths are sublinear, \Cref{th: av-D-OGD} ensure that the generalization ability of the output of D-OGD is increasing through time. This is informative on the robustness to the intrinsic randomness of the learning problem.
Note that our result holds for any sequence $\xbf$ such that $\xbf_t$ is $\mathcal{F}_{t-1}$-measurable. We present in \Cref{sec: toy_experiment} a toy experiment that exploits this additional flexibility by showing not only that it may not be relevant to compare ourselves to the true minimizers $\xbf^*$, but also that the optimistically tempered approach outperforms both OGD and OMGD.

\section{Experiments}
\label{sec: experiments}
This section aims to compare the efficiency of Optimistically Tempered approach compared to classical methods. We show here it leads to comparable or enhanced numerical results.
We propose two sets of experiments. The first one gathers 4 classical datasets two regression and two classification problems. Its goal is to assess our algorithm's efficiency by plotting the averaged cumulative losses $\sum_{i=1}^{t} \ell\left(h_{i},z_i\right) / t$ at any time $t$. The second experiment is a toy example designed to show that D-C-Risk is a relevant tool to handle learning processes on noisy problems.
For those two experiments, we compute three algorithms: the celebrated Online Gradient Descent \citep[][Alg. 1]{zinkevich2003online}, the D-OGD algorithm (\Cref{alg: dynamic_OGD}) and a variant of the Online Multiple Gradient Descent (OMGD) algorithm with decreasing steps \citep[][Alg. 1]{zhang2017improved}.
\\
The reason we computed OMGD is that \textsc{Construct} (\Cref{alg: additional_knowledge_OMGD}) is following the same idea as OMGD (\emph{i.e.}, performing a gradient descent at each time step for more accurate predictors). An interesting question is whether D-OGD provides similar or better results than OMGD. We address this below.
Furthermore, we would expect that using the output of \textsc{Construct} as experts instead of predictor would provide us additional flexibility in our learning process, is it the case in practice?

\subsection{Experiments on real-life datasets}
\label{sec: real-life-exp}
We conduct experiments on a few real-life datasets, in classification and regression. Our objective is twofold: check the convergence of our learning methods and compare their efficiencies with classical algorithms.
\\
\textbf{Binary Classification.} At each round $t$ the learner receives a data point $x_{t} \in \mathbb{R}^{d}$ and predicts its label $y_{t} \in\{-1,+1\}$ using $\left\langle x_{t}, \hat{\xbf}_{t}\right\rangle$, with $\xbf_t$ being the predictor given by the online algorithm of interest.
The adversary reveals the true value $y_{t}$, then the learner suffers the loss $\ell(\xbf_t,z_t)=\left(1-y_{t} \hat{\xbf}_{t}^{T} x_{t}\right)_{+}$ with $z_t=(x_t,y_t)$ and $a_{+}=a$ if $a>0$ and $a_{+}=0$ otherwise.
\\
\textbf{Linear Regression.} At each round $t$, the learner receives a set of features $x_{t} \in \mathbb{R}^{d}$ and predicts $y_{t} \in \mathbb{R}$ using $\left\langle x_{t}, \hat{\xbf}_{t}\right\rangle$ with $\xbf_t$ being the predictor given by the online algorithm of interest.
Then the adversary reveals the true value $y_{t}$ and the learner suffers the loss $\ell(\xbf_t,z_t)=\left(y_{t}-\hat{\xbf}_{t}^{T} x_{t}\right)^{2}$ with $z_t=(x_t,y_t)$.
\\
\textbf{Datasets.} We consider four real-world datasets: two for classification (Breast Cancer and Pima Indians), and two for regression (Boston Housing and California Housing). All datasets except the Pima Indians have been directly extracted from \texttt{sklearn} \citep{scikit-learn}.
Breast Cancer dataset \citep{street1993nuclear} is available \href{https://archive.ics.uci.edu/ml/datasets/Breast+Cancer+Wisconsin+(Diagnostic)}{here} and comes from the UCI ML repository as well as the Boston Housing dataset \citep{belsley2005regression} which can be obtained \href{https://archive.ics.uci.edu/ml/machine-learning-databases/housing/}{here}. California Housing dataset \citep{pace1997sparse} comes from the StatLib repository and is available \href{https://www.dcc.fc.up.pt/~ltorgo/Regression/cal_housing.html}{here}.
Finally, Pima Indians dataset \citep{smith1988using} has been recovered from this Kaggle \href{https://www.kaggle.com/datasets/uciml/pima-indians-diabetes-database}{repository}.  Note that we randomly permuted the observations to avoid learning irrelevant human ordering of data (such as date or label).
\\
\textbf{Parameter settings.}
 We ran our experiments on a 2021 MacBookPro with an M1 chip and 16 Gb RAM all our experiments are averaged over 20 runs of the algorithms.
 For OGD, the initialisation point is $\mathbf{0}_{\mathbb{R}^d}$ and the values of the learning rates are set to $\eta=1 / 2\sqrt{m}$. where $m$ is the size of the considered dataset.
 For OMGD, we ran the procedure while, at time $t$, performing a gradient descent with $K= 100$ iterations. This auxiliary gradient descent has been performed with steps $(\lambda/2\sqrt{j})_{j=1\cdots K}$. $\lambda$ ,being an empirical stabiliser set to $0.1/\sqrt{m}$.
 For D-OGD, we ran the procedure with a constant step $\eta= 0.1/\sqrt{m}$. We ran \textsc{Construct} to generate our experts with the iteration number $K=100$ and steps $(\eta'_j)_{j=1\cdots K}=(\lambda/2\sqrt{j})_{j=1\cdots K}$, $\lambda$ ,being an empirical stabiliser set to $0.1/\sqrt{m}$.

\begin{figure}
   \centering
   \begin{subfigure}[b]{0.3\textwidth}
     \centering
     \includegraphics[width=\textwidth]{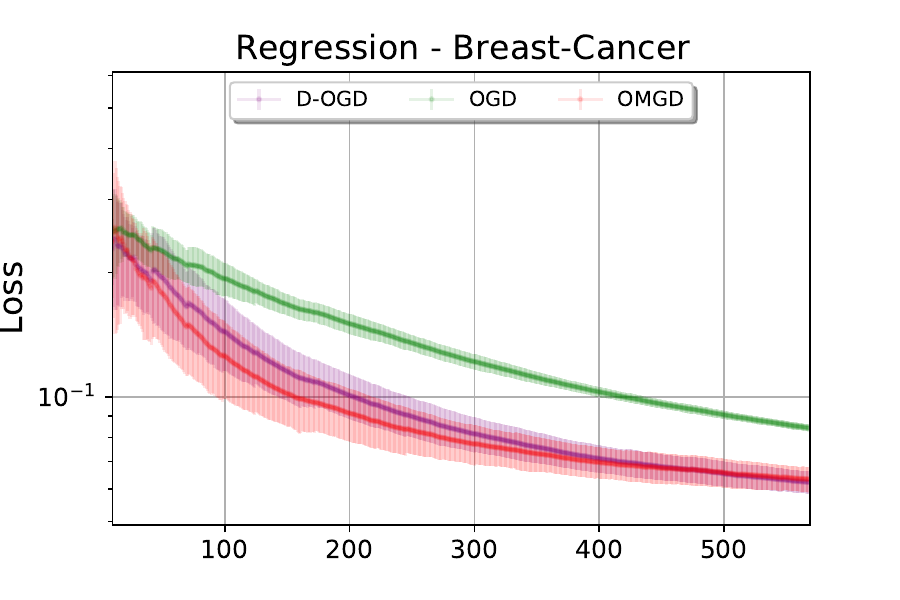}
  \end{subfigure}~
  \begin{subfigure}[b]{0.3\textwidth}
    \centering
    \includegraphics[width=\textwidth]{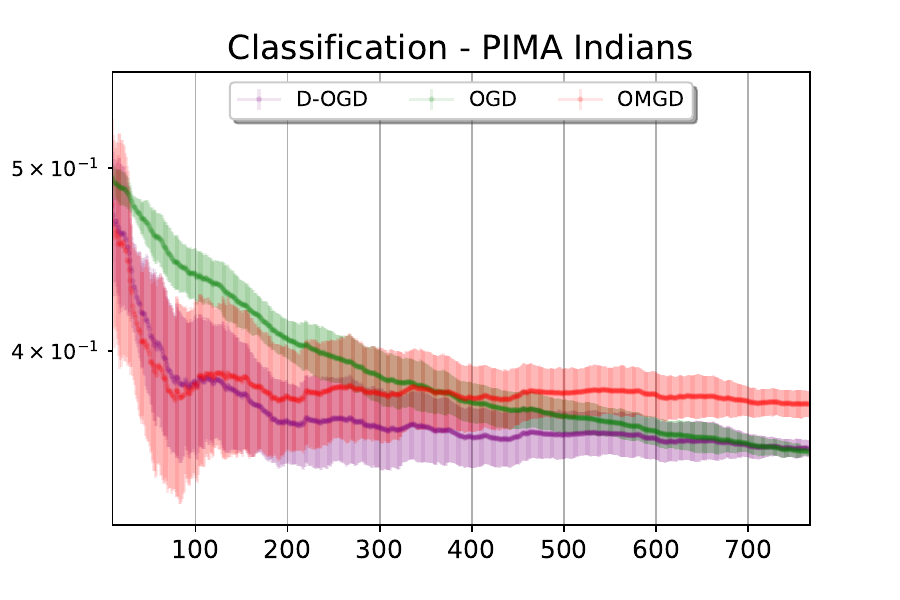}
 \end{subfigure}
 \begin{subfigure}[b]{0.3\textwidth}
   \centering
   \includegraphics[width=\textwidth]{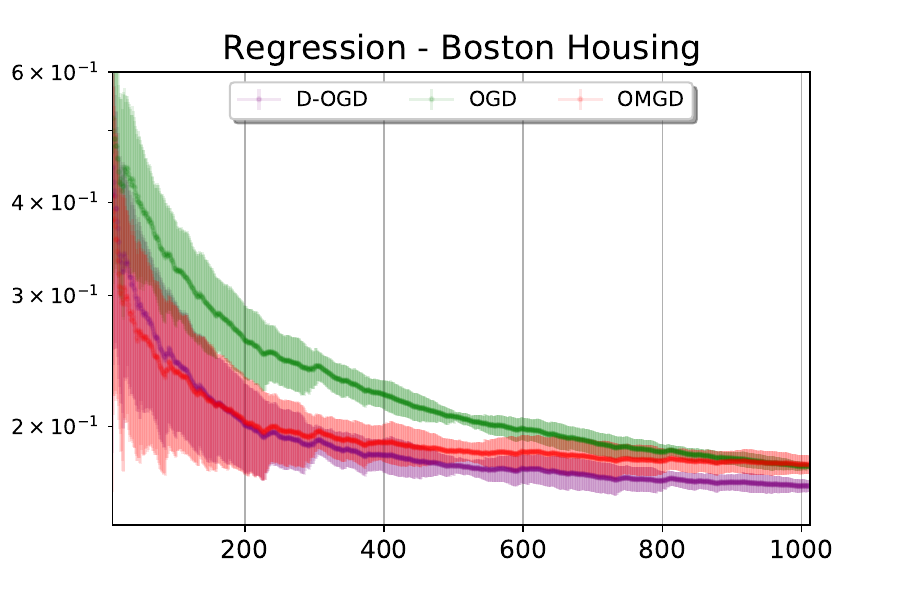}
\end{subfigure}~
\begin{subfigure}[b]{0.3\textwidth}
  \centering
  \includegraphics[width=\textwidth]{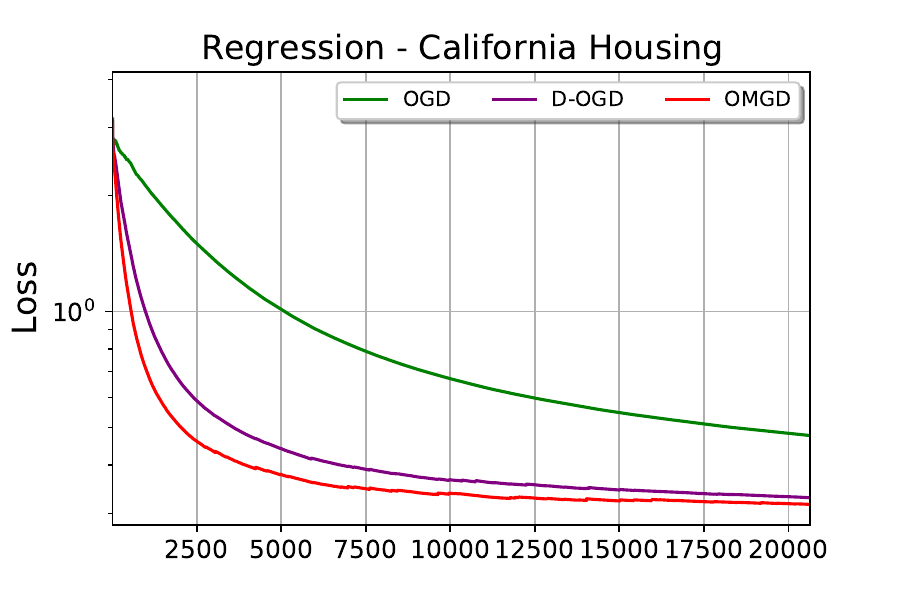}
\end{subfigure}
   \caption{Averaged cumulative losses for all datasets with error bars over 20 runs. The $x$-axis is the time.}
   \label{fig: real_datasets}
 \end{figure}

 \textbf{Quantity of interest.}
 For each dataset, we plot the evolution of the averaged cumulative loss $\sum_{i=1}^{t} \ell\left(h_{i},z_i\right) / t$ as a function of the step $t=1,..., m$, where $m$ is the dataset size and $h_{i}$ is the decision made by the learner $h_i$ at step $i$. The results are gathered in \Cref{fig: real_datasets} and are averaged over 20 runs with associated error bars.

\textbf{Empirical findings.} On those datasets, OMGD with adaptive steps and D-OGD seem to perform rather equivalently, except on the PIMA Indians dataset where D-OGD outperforms OMGD. On two datasets (Breast Cancer and California Housing), D-OGD performs better than OGD, otherwise, both methods perform similarly. A reason that could explain the efficiency of our method compared to OMGD in the PIMA Indians dataset is that because this problem is difficult (\emph{i.e.,} noisy), the technical condition stated in \cite[][Corollary 4]{zhang2017improved} may not be satisfied. This would impeach OMGD to attain competitive results.
Furthermore, note that in any case, D-OGD is at least as good as OGD or OMGD. The take-home message is that the OT-OL approach is comparably efficient on those datasets.

\subsection{A toy experiment: the Online Quadratic Problem}
\label{sec: toy_experiment}

\textbf{Theoretical framework.}
Our problem is set as follows: at each time step $t$, a random variable $\theta_t$ is drawn. For all $t$, $\theta_t$ is such that
\[P_t=\mathcal{L}(\theta_t\mid \mathcal{F}_{t-1})= \mathcal{N}(\texttt{moy}_t,\sigma_t^2).\]
We assume that there exists $D_m,D_\sigma$ positive values such that for all $t$, $(\texttt{moy}_t,\sigma_t) \in [-D_m/2,D_m/2]\times [0;D_\sigma]$.
Finally, we consider the losses
$\ell_t(\theta) = (\theta_t- \theta)^2$.
We refer to this framework as the \emph{Online Quadratic Problem}.

\textbf{Quantity of interest.}
We study the D-C-Risk w.r.t. the sequence $\xbf_t= \texttt{moy}_t$.
We cannot compare ourselves to the true minimizer $\xbf_t^*= \theta_t$ because this quantity is not $\mathcal{F}_{t-1}$ measurable. However, we show below that there exists another meaningful comparator.
Indeed, in our setup, we note that $\texttt{moy}_t$ was assumed to be $\mathcal{F}_{t-1}$-measurable so let us see what gives the dynamic cumulative risk for any sequence of predictors $(\hat{\xbf}_t)_{t\geq 0}$:
  $\sum_{t=1}^T L_t(\hat{\xbf}_t)- \sum_{t=1}^T L_t(\texttt{moy}_t)  = \sum_{t=1}^T\mathbb{E}_{t-1}[(\theta_t-\hat{\xbf}_t)^2] - \sum_{t=1}^T\mathbb{E}_{t-1}[(\theta_t-\texttt{moy}_t)^2]  = \sum_{t=1}^T (\hat{\xbf}_t- \texttt{moy}_t)^2.$

The last line holding thanks to a bias-variance tradeoff, this basic calculation shows that for this learning problem, using $(\texttt{moy}_t)_t$ as comparators instead of the true minimizers leads to a meaningful regret. Yet, we can derive from the general notion of dynamic regret a comparison between our prediction and the true mean of the data. One will see in the experiments that D-OGD can approximate the means better than classical OGD at high times.

\textbf{Parameter settings.}
All our algorithms are using a projection on the ball centered in $0$ of diameter $D=10$.
For OGD, the initialisation point is $\mathbf{0}_{\mathbb{R}^d}$ and the values of the learning rates are set to $\eta_t =1 / 2\sqrt{t}$.
For OMGD, we ran the procedure while, at time $t$, performing a gradient descent with $K= 100$ iterations. This auxiliary gradient descent has been performed at time $t$ with steps $(\lambda_t/2\sqrt{j})_{j=1\cdots K}$, $\lambda_t$ being an empirical stabiliser set to $1/2\sqrt{t}$.
For D-OGD, we ran two variants: the first uses \textsc{Construct} to generate our experts. We run \cref{alg: dynamic_OGD} with steps $\eta_t= 1/2\sqrt{t}$ at time $t$. We run \textsc{Construct} with, at each time $t$, the iteration number $K=100$ and steps $(\eta'_j)_{j=1\cdots K}=(\lambda_t/2\sqrt{j})_{j=1\cdots K}$, $\lambda_t$ being an empirical stabiliser set to $1/2\sqrt{t}$.
The second does not use \textsc{Construct} and instead defines at each time $t$ $\nu_{t}\sim \mathcal{N}(\hat{\xbf}_t,\sigma_1^2)$ with $\sigma_1 = 0.4$. Similarly, we run \cref{alg: dynamic_OGD} with steps $\eta_t= 1/2\sqrt{t}$ at time $t$.

\begin{wrapfigure}{r}{0.5\textwidth} % r=right, l=left
    \centering
    \vspace{-15pt} % shift up if needed
    \includegraphics[width=0.4\textwidth]{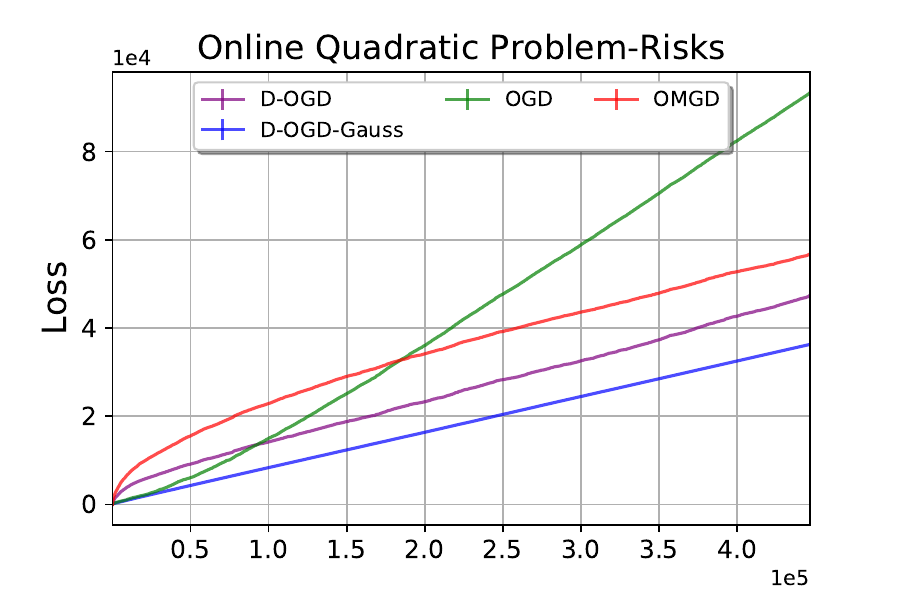}
    \caption{Cumulative risks of D-OGD (purple,blue), OMGD (red), OGD (green).}
    \label{fig: toy_experiment}
\end{wrapfigure}

\textbf{Experimental framework.}
We take for any $t$, $\texttt{moy}_t= \sin\left( \frac{t}{\omega} \right)$ with $\omega=200$, yet the means are a deterministic sequence fixed before our study. Then our $\theta_t$ are drawn independently. We also fix for any $t$, $\sigma_t=\sigma=4$. We chose $K$ (the number of iterations to acquire our experts) equal to $100$. Results are gathered in \Cref{fig: toy_experiment}.

% \begin{figure}[!h]
%   \centering
%     \includegraphics[scale= 0.5]{../Code/Toy_experiment/online_quad_prob.pdf}
%   \caption{Cumulative risks of D-OGD (purple,blue), OMGD (red), OGD (green).  }
%   \label{fig: toy_experiment}
% \end{figure}

\textbf{Empirical findings.}
First, OGD fails on this example as the problem is too noisy: OGD fails to detect any statistical pattern between the successive points.
Second, OMGD performs better than OGD but is significantly worse than D-OGD (the difference of the dynamic cumulative risks is of magnitude $10^4$). This shows that our method, which only uses the output of the auxiliary gradient descent as experts (and not as predictors as in OMGD) provides flexibility that translates into a greater performance for extremely noisy problems. A reason that could explain the efficiency of our method compared to OMGD is again that the intrinsic noise is so high that the technical condition stated in \cite[][Cor. 4]{zhang2017improved} may not be satisfied, which impeaches OMGD to attain a competitive dynamic regret in $\mathcal{O}(\min (P_T^*,S_T^*))$.
Finally, note that interestingly, our variant of D-OGD (the curve 'DOGD Gauss' which uses an alternative source of additional information) provides better results here while we have no theoretical guarantee of its efficiency. This opens the way to a broader reflection on the choice of experts within D-OGD.
In conclusion, this experiment shows that the OT-OL approach outperforms both OGD and OMGD: exhibiting the interest of treating expert advice with caution instead and granting them full trust.

\section{Conclusion}

We introduced the notion of Tempered Optimism in Online Learning as well as various original algorithms for both non-convex and convex scenarios. Our non-convex algorithms (\Cref{alg: OT_OGD,alg: OT-OMD}) enjoys dimension-free regret bounds, a significant improvement compared to the state-of-the-art in Online Non-Convex Learning, at the cost of an approximate optimisation oracle. To avoid such an oracle, our algorithm for convex losses (\Cref{alg: dynamic_OGD}) uses explicitly crafted experts, whose quality is controlled thanks to convexity to get sound dynamic regret bounds. In both cases, we required experts to be good approximations of (local) minima. This raises two questions: \textit{(i)}, if we maintain this property, is \textsc{Construct} (\Cref{alg: additional_knowledge_OMGD}) the best way to do so? Indeed, it is not the only choice as we could use, \emph{e.g.}, the Newton algorithm for learning problems in small dimensions as this algorithm is known to converge quickly. This leverages an experimental tradeoff between accuracy and time complexity involving the dimension as a hyperparameter of the problem. \textit{(ii)} Can we identify alternative properties that we want our expert to satisfy (and identify the associated limitations) in order to exploit further the flexibility of Tempered Optimism?
% \\
% Another promising lead lies in the flexibility of the OT-OL framework when choosing expert advice, as we can confidently propose novel types of advice knowing they will be ignored by \textsc{Adjust} if useless. More precisely, in this work, $\nu$ focuses on being a good approximation of the minima sequence while our bounds involve a broader tradeoff between path lengths (\emph{i.e.,} only small shifts are recommended through time) and being a good approximation of the past minimizers. 

\bibliography{biblio}
\newpage

%\newpage
\appendix

\section*{Structure of the appendix}
The appendix is structured as follows: we start with additional background (\Cref{sec: technical_b}), further details on our procedures (\cref{sec: inspiration_perf,sec: construct}), additional OT algorithms for convex losses in \Cref{sec: additional-OT-algos}, and we defer unstated proofs to \cref{sec: proof-nonconvex,sec: proof_determin_results,sec: proof_proba_results}.

\section{Technical background}
\label{sec: technical_b}

\subsection{Azuma-Hoeffding's inequality}

One recalls the celebrated Azuma- Hoeffding inequality

\begin{proposition}
  \label{prop: Azuma_Hoeffding}
  Let $\left\{X_{0}, X_{1}, \cdots\right\}$ be a martingale with respect to filtration $\left\{\mathcal{F}_{0}, \mathcal{F}_{1}, \cdots\right\}$. Assume there are predictable processes $\left\{A_{0}, A_{1}, \cdots\right\}$ and $\left\{B_{0}, B_{1}, \ldots\right\}$ with respect to $\left\{\mathcal{F}_{0}, \mathcal{F}_{1}, \cdots\right\}$, i.e. for all $t, A_{t}, B_{t}$ are $\mathcal{F}_{t-1}$-measurable, and constants $0<c_{1}, c_{2}, \cdots<\infty$ such that
  $$
  A_{t} \leq X_{t}-X_{t-1} \leq B_{t} \quad \text { and } \quad B_{t}-A_{t} \leq c_{t}
  $$
  almost surely. Then for all $\epsilon>0$,
  $$
  \mathrm{P}\left(\left|X_{n}-X_{0}\right| \geq \epsilon\right) \leq 2 \exp \left(-\frac{2 \epsilon^{2}}{\sum_{t=1}^{n} c_{t}^{2}}\right)
  $$
\end{proposition}

In this work we use Azuma-Hoeffding's bound in the particular case where $A_t,B_t$ are constants almost surely.

\subsection{Bregman divergences}

We recall some background from \cite{hazan2019introduction}. 

Consider regularisation functions, denoted $R: \mathcal{K} \mapsto \mathbb{R}$, which are $\alpha$-strongly convex $\beta$-smooth, and $G_R$-Lipschitz.
Also assume that $R$ is twice differentiable over $\mathcal{K}$. Then, by strong convexity, for all $x \in \operatorname{int}(K)$, the Hessian $\nabla^2 R(\xbf)$ is positive definite.

In the following we make use of norms $\|.\|_H$ associated to the scalar product $\langle \xbf,\ybf\rangle_H := \xbf^{\top}H\ybf$, where $A$ is positive definite, and their dual. The dual norm associated to $\|.\|_H$ is defined as $\|.\|_H^* := \|.\|_{H^{-1}}$.

We consider matrix norms with respect to $\nabla^2 R(\ybf)$, as well as the inverse Hessian $\nabla^{-2} R(\ybf)$. In such cases, we use the notations $\|\xbf\|_{\ybf}:=\|\xbf\|_{\nabla^2 R(\ybf)}$ and $
\|\xbf\|_{\ybf} :=\|\xbf\|_{\nabla^{-2} R(\ybf)}$.

\begin{definition}
Denote by $B_R(\xbf \mid \ybf)$ the Bregman divergence with regularization $R$ is defined as
\[B_R(\xbf \mid \ybf)=R(\xbf)-R(\ybf)-\nabla R(\ybf)^{\top}(\xbf-\ybf) .\]
\end{definition} 
 
Recall that, by Taylor expansion and the mean-value theorem gives the following formulation of Bregman divergence:
$B_R(\xbf \mid \ybf)=\frac{1}{2} \| \xbf-\ybf \|_{\z}^2
$, for some $\z = v \xbf + (1-v)\ybf, v\in[0,1]$ \citep{hazan2019introduction}.
Therefore, the Bregman divergence defines a local norm, which has a dual norm, denoted respectively by $\|.\|_{\xbf,\ybf} := \|.\|_{\z}, \|.\|_{\xbf,\ybf}^{*} :=  \|.\|_{\z}^*$.
Then, we can write

$$
B_R(\xbf \mid \ybf)=\frac{1}{2} \| \xbf-\ybf \|_{\xbf, \ybf}^2 .
$$

\section{Inspiration for our notion of performance in \Cref{sec: convex}}
\label{sec: inspiration_perf}

\noindent Let $\mathbf{\eta}= (\eta_t)_{t=1..T)}$ be a positive step sequence.

\noindent We denote by $\hat{\xbf}_t, t\geq 1$ the sequence of predictors defined by the classical projected OGD:

\[ \hat{\xbf}_{t+1} = \Pi_{\mathcal{K}}\left( \hat{\xbf}_t - \nabla \ell_t(\hat{\xbf}_t)   \right)    \]

    \begin{theorem}
        \noindent Dynamic regret of projected OGD on a closed convex $\mathcal{K}$ for convex losses with steps $\mathbf{\eta}= (\eta_t)_{t=1..T)}$ with regards to $\mathbf{\xbf}=(\xbf_t)_{t=0..T}\in\mathcal{K}^T$ satisfies :

        \begin{align*}
            \sum_{t=1}^T \ell_t(\hat{\xbf}_t) - \sum_{t=1}^T \ell_t(\xbf_t) & \leq \sum_{t=1}^T \langle\nabla(\ell_t), \hat{\xbf}_t- \xbf_t\rangle \\
            & \leq \frac{D^2}{2\eta_T} + \frac{G^2}{2}\sum_{t=1}^T\eta_t  -\sum_{t=1}^T \frac{\mathrm{Perf}(t,\hat{\xbf},\xbf)}{\eta_t}.
        \end{align*}

    \end{theorem}

\begin{proof}

First, convexity of the losses gives us :
$$
\sum_{t=1}^{T} \ell_{t}\left(\hat{\xbf}_{t}\right)-\sum_{t=1}^{T} \ell_{t}\left(\xbf_{t}\right) \leqslant \sum_{t=1}^{T}\left\langle\nabla \ell_{t}\left(\hat{\xbf}_{t}\right), \hat{\xbf}_{t}-\xbf_{t}\right\rangle
$$

\noindent To control the right hand side of this bound we use:
$$
\begin{aligned}
\left\|\hat{\xbf}_{t+1}-\xbf_{t}\right\|^{2} & \leqslant\left\|\hat{\xbf}_{t}-\eta_{t} \nabla \ell_{t}\left(\hat{\xbf}_{t}\right)-\xbf_{t}\right\|^{2} \\
&=\left\|\hat{\xbf}_{t}-\xbf_{t}\right\|^{2}-2 \eta_{t}\left\langle\nabla \ell_{t}\left(\hat{\xbf}_{t}\right), \hat{\xbf}_{t}-\xbf_{t}\right\rangle+\eta_{t}^{2}\left\|\nabla \ell_{t}\left(\hat{\xbf}_{t}\right)\right\|^{2}
\end{aligned}
$$
\noindent Hence: $$\left\|\hat{\xbf}_{t+1}-\xbf_{t+1} \right\|^{2} \leqslant\left\| \hat{\xbf}_{t}-\xbf_{t} \right\|^{2}-2 \eta_{t}\left\langle\nabla \ell_{t}\left(\hat{\xbf}_{t}\right), \hat{\xbf}_{t}-\xbf_{t}\right\rangle+\eta_{t}^{2} G^{2}-2 \mathrm{Perf}\left(t, \hat{\xbf},\xbf)\right.$$

\noindent So:

$$\left\langle\nabla \ell_{t}\left(\hat{\xbf}_{t}\right), \hat{\xbf}_{t}-\xbf_{t}\right\rangle \leqslant \frac{\left\|\hat{\xbf}_{t}-\xbf_{t}\right\|^{2}-\left\|\hat{\xbf}_{t+1}- \xbf_{t+1}\right\|^{2}}{2 \eta_{t}}+\frac{\eta_{t} G^{2}}{2}-\frac{\mathrm{Perf}(t, \hat{\xbf}, \xbf)}{\eta_{t}}$$

Summing on $t$ gives (assuming $1/\eta_0=0$):

\begin{align*}
  \sum_{t=1}^{T}\left\langle\nabla \ell_{t}\left(\hat{\xbf}_{t}\right), \hat{\xbf}_{t}-\xbf_{t}\right\rangle & \leq \sum_{t=1}^T ||\hat{\xbf}_t - \xbf_t||^2 \left( \frac{1}{2\eta_t} -  \frac{1}{2\eta_{t-1}} \right)
  +\frac{G^{2}}{2}\sum_{t=1}^T \eta_{t} -\sum_{t=1}^T\frac{\mathrm{Perf}(t, \hat{\xbf}, \xbf)}{\eta_{t}}\\
  & \leq D^2 \sum_{t=1}^T \left( \frac{1}{2\eta_t} -  \frac{1}{2\eta_{t-1}} \right)
  +\frac{G^{2}}{2}\sum_{t=1}^T \eta_{t} -\sum_{t=1}^T\frac{\mathrm{Perf}(t, \hat{\xbf}, \xbf)}{\eta_{t}}\\
  & \leq \frac{D^2}{\eta_T}  +\frac{G^{2}}{2}\sum_{t=1}^T \eta_{t} -\sum_{t=1}^T\frac{\mathrm{Perf}(t, \hat{\xbf}, \xbf)}{\eta_{t}}
\end{align*}

\end{proof}

\noindent One can also have a stronger result for $\lambda$-strongly convex functions with the following additional assumption:

\noindent We assume that our steps $\eta_t$ are such that:
    \[ \frac{1}{\eta_t}-\lambda \leq \frac{1}{\eta_{t-1}}  \]
    \begin{theorem}

    \noindent Dynamic regret of projected OGD on a closed convex $\mathcal{K}$ with steps $\mathbf{\eta}= (\eta_t)_{t=1..T)}$ with regards to $\mathbf{\xbf}=(\xbf_t)_{t=0..T}\in\mathcal{K}^T$ satisfies :

        \begin{align*}
            \sum_{t=1}^T \ell_t(\hat{\xbf}_t) - \sum_{t=1}^T \ell_t(\xbf_t)
            & \leq  \frac{G^2}{2}\sum_{t=1}^T\eta_t -\sum_{t=1}^T \frac{\mathrm{Perf}(t,\hat{\xbf},\xbf)}{\eta_t}.
        \end{align*}

    \end{theorem}

\begin{proof}
The proof is roughly the same than the one for the previous bound. We remark that thanks to strong convexity, one now has :

$$
\sum_{t=1}^{T} \ell_{t}\left(\hat{\xbf}_{t}\right)-\sum_{t=1}^{T} \ell_{t}\left(\xbf_{t}\right) \leqslant \sum_{t=1}^{T}\left\langle\nabla \ell_{t}\left(\hat{\xbf}_{t}\right), \hat{\xbf}_{t}-\xbf_{t}\right\rangle - \lambda ||\hat{\xbf}_t-\xbf_t||^2
$$

So the arguments of the previous proof provide us:

\begin{align*}
  \sum_{t=1}^T \ell_t(\hat{\xbf}_t) - \sum_{t=1}^T \ell_t(\xbf_t)  & \leq \frac{1}{2}\sum_{t=1}^T \left(   \frac{1}{ \eta_{t}} - \lambda \right)\left\| \hat{\xbf}_{t}-\xbf_{t}\right\|^{2}  - \frac{\left\|\hat{\xbf}_{t+1}-\hat{\xbf}_{t+1}\right\|^{2}}{\eta_t} \\
  & +  \sum_{t=1}^T\frac{\eta_{t} \left\|\nabla \ell_{t}\left(\hat{\xbf}_{t}\right)\right\|^{2}}{2}-\frac{\mathrm{Perf}(t, \hat{\xbf}, \xbf)}{\eta_{t}}\\
  & \leq \frac{1}{2}\sum_{t=1}^T  \frac{ \left\| \hat{\xbf}_{t}-\xbf_{t}\right\|^{2}}{\eta_{t-1}}  - \frac{\left\|\hat{\xbf}_{t+1}-\hat{\xbf}_{t+1}\right\|^{2}}{\eta_t} \\
  & +  \sum_{t=1}^T\frac{\eta_{t} \left\|\nabla \ell_{t}\left(\hat{\xbf}_{t}\right)\right\|^{2}}{2}-\frac{\mathrm{Perf}(t, \hat{\xbf}, \xbf)}{\eta_{t}}
\end{align*}

A telescopic argument and bound over the gradients provides us the final result.

\end{proof}

\begin{remark}
We focus in three specific cases where performance can be linked to classical quantities:
    \begin{itemize}
        \item First is just a remark : we totally recover the classical OGD bound for static regret when one has $\xbf_{t+1}=\xbf_t$ for any $t$.
        \item Second, if our OGD predicts well the minimiser $\xbf^*$ after a certain time, i.e. for $t\geq t_0$, $\hat{\xbf}_{t+1}\approx \xbf_{t+1}^*$. Then one has
        \[ \sum_{t=1}^T\mathrm{Perf}(t,\hat{\xbf}, \xbf) \approx - \frac{1}{2}\sum_{t=1}^T \frac{||\xbf_{t+1}^* - \xbf_t^*||^2}{\eta_t} \leq -\frac{1}{\eta_1}S^*_T.  \]
        so our result ensures that in this case, OGD has been able to tame the geometry induced by the $\ell_t$s to generate a momentum greater than $S^*_T/\eta_1$

        \item Finally let us consider the overfitting case i.e, for each $t$, $\hat{\xbf}_{t+1}\approx \xbf^*_t$.
        Then:
        \[ \sum_{t=1}^T\mathrm{Perf}(t,\hat{\xbf}, \xbf) \approx -\frac{1}{2}\sum_{t=1}^T \frac{||\xbf_{t+1}^* - \xbf_t^*||^2}{\eta_t} \leq \frac{1}{\eta_T}S^*_T.  \]

        \noindent So overfitting will penalise our OGD with at most a factor $\mathcal{S}_T/\eta_T$

    \end{itemize}
\end{remark}

However, even if our bounds gives us an intuition on how is the OGD interacting with its environment. One cannot control it directly. If we assume having additional information at each time steps, this notion of performance can help us to enhance OGD.

\section{An explicit construction of experts in \Cref{sec: convex}}
\label{sec: construct}
As mentioned in \Cref{sec: convex}, we craft our experts by multiple gradient descent steps. We provide an algorithmic statement of this claim in \Cref{alg: additional_knowledge_OMGD} and also recall a convergence property of Gradient Descent to its optimum.

\begin{algorithm}
 \SetAlgoLined
 \SetKwInOut{Initialisation}{Initialisation}
 \SetKwInOut{Parameter}{Parameters}
 \Parameter{The number $K$ of iterations, step-sizes $(\eta_j')_{j=1..K}$\\
 Current loss function $\ell_t$, current point $\hat{\xbf}_t$}
 \Initialisation{Set $\mathbf{y}_{0}:= \hat{\xbf}_t$}
\textbf{For $j$ in $0..K-1$:}\\
Update \[ \mathbf{y}_{j+1} = \Pi_{\mathcal{K}}\left( \mathbf{y}_{j} -  \eta_j' \nabla \ell_t(\mathbf{y}_{j}) \right) \] \\
\textbf{Return $\nu_{t}:=\frac{1}{K}\sum_{j=1}^{K} \mathbf{y}_{j}$}
 \caption{The \textsc{Construct} algorithm.}
 \label{alg: additional_knowledge_OMGD}
 \end{algorithm}

  We recall in \cref{l: GD_add_know} a convergence property of the gradient descent algorithm.

\begin{lemma}
  \label{l: GD_add_know}
  Assume $\ell_t$ to be $\lambda$-strongly convex and that $(\eta'_j)$ verify for all $j$, $\frac{1}{\eta'_j} - \lambda \leq \frac{1}{\eta'_{j-1}}$.
  Then for any $t$ we have, 
  \[ \ell_t(\nu_{t}) - \inf_{\xbf \in \mathcal{K}}\ell_t(\xbf) \leq \frac{G^2}{K} \sum_{j=1}^K  \eta'_j.\]

  If losses are only convex and $\eta'_j= \frac{D}{G\sqrt{j}}$ for all $j$, we then have: 
  \[ \ell_t(\nu_{t}) - \inf_{\xbf \in \mathcal{K}}\ell_t(\xbf) \leq \frac{3}{2\sqrt{K}}GD. \]
\end{lemma}
 Remark that it is essential to consider strongly convex functions to obtain the rate of \cref{l: GD_add_know}. To satisfy the technical condition on the step sizes, we can consider the step sequence $(\frac{1}{\lambda t^\alpha})_{t\geq 1}$ for any $\alpha\in [0,1]$.

 \begin{proof}
  Let $t\geq 0$ and $\xbf_t^*$ minimising $\ell_t$. Recall that $\nu_{t}$ is defined as the Polyak averaging $\nu_{t}:=\frac{1}{K}\sum_{j=1}^{K} \ybf_{j}$. First, we remark that by convexity of $\ell_t$:

  \begin{align*}
      \ell_t(\nu_{t}) - \ell_t(\xbf_t^*) & = \ell_t\left(\frac{1}{K}\sum_{j=1}^K \ybf_{j} \right)- \ell_t(\xbf_t^*) \leq \frac{1}{K}\sum_{j=1}^K \ell_t(\ybf_j) -\ell_t(\xbf_t^*).
  \end{align*}

    Because \textsc{Construct} is a gradient descent with steps $(\eta'_j)_{j=1..K}$ on the $\lambda$-strongly convex function $\ell_t$, one has for any $j$,
  the classical route of proof for static regret bound for strongly convex functions described in \cite[Theorem 3.3]{hazan2019introduction}. One then has the following, which concludes the proof:
\begin{align*}
  \sum_{j=0}^K \left(\ell_t(\ybf_j) -\ell_t(\xbf_t^*)\right) & \leq G^2\sum_{j=1}^K \eta_j'.
\end{align*}

If losses $\ell_t$ are simply convex, we then apply the classical OGD proof of \citet[Theorem 3.1]{hazan2019introduction} to conclude the proof.
\end{proof}

\section{Supplementary Optimistically Tempered algorithms for strongly convex losses}
\label{sec: additional-OT-algos}

In this section, we show that \textsc{Adjust} is flexible enough to provide Optimistically Tempered adaptations of Online Newton Step \citep{hazan2007logarithmic} and Adagrad \cite{duchi2011adaptive}, extending the range of application of Tempered Optimism far beyond OGD. To do so, we assume that for all $t$, $\ell_t$ is $\lambda$-strongly convex: 

\[ \forall (t, \xbf,\xbf_0 )\in \mathbb{N}/\{0\}\times \mathcal{K}^2, \ell_t(\xbf) - \ell_t(\xbf_0) \leq \langle \nabla \ell_t(\xbf), \xbf- \xbf_0 \rangle - \lambda \|\xbf-\xbf_0\|^2.   \]

We are then able to provide our novel algorithms.

\subsection{Optimistically tempered Online Newton Step}

\Cref{alg: dynamic_ONS} details the OT-ONS algorithm, which is an optimistically tempered version of ONS \citep{hazan2007logarithmic}. We present in \cref{th: final_bound_DONS} its associated D-Regret bound.

\begin{algorithm}[!ht]
\SetAlgoLined
\SetKwInOut{Initialisation}{Initialisation}
\SetKwInOut{Parameter}{Parameters}
\Parameter{Horizon $T$, step $\gamma,\varepsilon>0$. }
\Initialisation{convex set $\mathcal{K}$, initial point $\xbf_1\in\mathcal{K}\subseteq \mathbb{R}^d$,additional information $\nu_{0}\in\mathcal{K}$,$A_{0}=\varepsilon I_d $ }
\textbf{For} $t$ in $\{1,\dots, T\}$: \\
\hspace{5mm} Update $A_{t}=A_{t-1}+\nabla_{t} \nabla_{t}^{\top}$\\
\hspace{5mm} Set $\hat{\xbf}_{temp,t+1} = \hat{\xbf}_{t} - \frac{1}{\gamma} A_{t}^{-1} \nabla_{t}$ \\
\hspace{5mm}Get $\nu_{t}$ \\
\hspace{5mm}$\hat{\xbf}_{t+1}= \textsc{Adjust}(t,A_t,\nu,\hat{\xbf}_{temp,t+1})$\\
\textbf{Return} $\hat{\xbf}=(\hat{\xbf}_{t})_{t=0..T}$
\caption{OT-ONS onto a closed convex space $\mathcal{K}$.}
\label{alg: dynamic_ONS}
\end{algorithm}

\begin{theorem}
    \label{th: final_bound_DONS}
    Denote by $\xbf_t^*= \operatorname{argmin}_{\xbf\in\mathcal{K}}\ell_t(\xbf)$. We assume that our predictors $\hat{\xbf}$ are obtained using OT-ONS (\Cref{alg: dynamic_ONS} ) with $\gamma= \frac{1}{2} \min \left\{\frac{1}{ G D}, \frac{\lambda}{G^2} \right\}, \; \varepsilon= \frac{1}{\gamma^2 D^2}  $.
    We also assume our experts $\nu$ to be the output of  \textsc{Construct} (\Cref{alg: additional_knowledge_OMGD}) used at time $t$ with steps $\mathbf{\eta'}=(\frac{1}{\lambda j})_{j=1..K}$ and $K=T$.
    Then, dynamic regret of OT-ONS with regards to  $\xbf^*= (\xbf_t^*)_{t=0..T}$ the true minimisers satisfy :
    \begin{align*}
       \sum_{t=1}^T \ell_t(\hat{\xbf}_t) - \sum_{t=1}^T \ell_t(\xbf_t^*) & \leq  GP_T(\nu) - \lambda S_T(\nu) +  2\left( \frac{G^2}{\lambda}(d+1) + dGD  \right)(1+ \log(T)).
    \end{align*}

    Furthermore, this result remains  for any additional knowlege $\nu$ such that for any $t$, $\ell_t(\nu_{t}) - \ell_t(\xbf_t^*) = \mathcal{O}(1/t)$.
\end{theorem}

\cref{th: final_bound_DONS} can be linked to the Online Multiple Newton Update (OMNU) when $\nu_{t}$ is the output of multiple Newton steps to approximate $\xbf_{t}^*$, one then can consider OT-ONS as an optimistically tempered version of OMNU. \citet{zhang2017improved} proposed a competitive rate of $\mathcal{O}(\min(P_T,S_T))$ for OMNU. While our rate is weaker than theirs, our results hold with the single assumption of strong convexity. 
Indeed, \citet[][Thm 11.]{zhang2017improved} holds under a set of technical assumptions \citet[][Assumption 10]{zhang2017improved} involving among others, the strict convexity of the losses and holding for problems having small variations of their successive minima. Our result requires fewer assumptions at the cost of $K=T$ iterations of \textsc{Construct} at each time step. As OT-ONS is an OT-OL algorithm it is expected to recover a deteriorated rate compared to OMNU which deals optimistically with experts.
Finally, taking $K_t= t$ at each time step allows us to not know in advance the stopping time of OT-ONS and recovers a slightly deteriorated rate of $\mathcal{O}(d\log(T)^2)$.

\begin{theorem}
    \label{th: av-OT-ONS}
    We assume that our predictors $\hat{\xbf}$ are obtained using OT-ONS(\Cref{alg: dynamic_ONS}) with $\gamma= \frac{1}{2} \min \left\{\frac{1}{ G D}, \frac{\lambda}{4G^2} \right\}, \; \varepsilon= \frac{1}{\gamma^2 D^2}  $.
    We also assume our experts $\nu$ to be the output of  \textsc{Construct} (\Cref{alg: additional_knowledge_OMGD}) used at time $t$ with steps $\mathbf{\eta'}=(\frac{1}{\lambda j})_{j=1..K}$ and $K=T$.
    Then, the dynamic cumulative risk satisfies with probability $1-2\delta$, for any $T \geq 1$,for any sequence $(\xbf_t)_{t=1..T}$ such that $\xbf_t$ is $\mathcal{F}_{t-1}$-measurable:
    \begin{align*}
        \sum_{t=1}^T L_t(\hat{\xbf}_t) - \sum_{t=1}^T L_t(\xbf_t)
        & \leq  GP_T(\nu) + 2G^2 S_T(\nu) + \mathcal{O}(d\log(T) + \log(1/\delta)),
    \end{align*}
    where $L_t= \mathbb{E}_{t-1}[\ell_t]$.
    This result remains  for any experts $\nu$ s.t. $\forall t, \ell_t(\nu_{t}) - \ell_t(\xbf_t^*) = \mathcal{O}(1/t)$.
\end{theorem}

\subsection{Optimistically tempered AdaGrad}
 \Cref{alg: dynamic_adagrad} details the OT-Adagrad algorithm, an optimistically tempered version of AdaGrad \citep{duchi2011adaptive} and we present in \Cref{th: final_bound_DAdagrad} its associated D-Regret bound. We use the notation $A \bullet B$ to denote the element-wise multiplication between the matrices $A$ and $B$.

\begin{algorithm}[!ht]
 \SetAlgoLined
 \SetKwInOut{Initialisation}{Initialisation}
 \SetKwInOut{Parameter}{Parameters}
 \Parameter{Horizon T, step $\eta$, parameter $\varepsilon$. }
 \Initialisation{Initial point $\xbf_1\in\mathcal{K}$,additional information $(\nu_{0})\in\mathcal{K}$,
 $G_{0}=\varepsilon \mathbf{I}_d,H_{0}=G_0^{1/2}$ }
\textbf{For} $t$ in $\{1,\dots, T\}$: \\
\hspace{5mm}Update $G_{t}=G_{t-1}+\nabla_{t} \nabla_{t}^{\top}$ \\
\hspace{5mm}Update $H_{t}=\underset{H \succeq 0}{\arg \min }\left\{G_{t} \bullet H^{-1}+\operatorname{Tr}(H)\right\}=G_{t}^{1 / 2}$ \\
\hspace{5mm}Set $\hat{\xbf}_{temp,t+1} = \hat{\xbf}_{t} -\eta H_{t}^{-1} \nabla_{t}$\\
\hspace{5mm}Get $\nu_{t}$\\
\hspace{5mm}$\hat{\xbf}_{t+1}= \textsc{Adjust}(t,H_t,\nu,\hat{\xbf}_{temp,t+1})$\\
\textbf{Return} $\hat{\xbf}=(\hat{\xbf}_{t})_{t=0..T}$
 \caption{OT-AdaGrad onto a closed convex space $\mathcal{K}$.}
 \label{alg: dynamic_adagrad}
 \end{algorithm}

\begin{theorem}
    \label{th: final_bound_DAdagrad}
    Denote by $\xbf_t^*= \operatorname{argmin}_{\xbf\in\mathcal{K}}\ell_t(\xbf)$. We assume that our predictors $\hat{\xbf}$ are obtained using OT-Adagrad (\Cref{alg: dynamic_adagrad} ) with $\varepsilon= \frac{2}{D^2}, \eta= \frac{D}{\sqrt{2}} $.
    We also assume our experts $\nu$ to be the output of  \textsc{Construct} (\Cref{alg: additional_knowledge_OMGD}) used at time $t$ with steps $\mathbf{\eta'}=(\frac{1}{\lambda j})_{j=1..K}$ and $K=T$.
    Then, dynamic regret of OT-Adagrad with regards to  $\xbf^*= (\xbf_t^*)_{t=0..T}$ the true minimisers satisfy :
    \begin{multline*}
       \sum_{t=1}^T \ell_t(\hat{\xbf}_t) - \sum_{t=1}^T  \ell_t(\xbf_t^*) \leq GP_T(\nu) - \lambda S_T(\nu)
       + \sqrt{2} D\left( 1 + \sqrt{\min_{H \in \mathcal{H}} \sum_{t}\left\|\nabla_{t}\right\|_{H}^{* 2}} \right) + \frac{G^2}{\lambda}(1+\log(T)).
    \end{multline*}

    Furthermore, this result remains  for any additional knowlege $\nu$ such that for any $t$, $\ell_t(\nu_{t}) - \ell_t(\xbf_t^*) = \mathcal{O}(1/t)$.

\end{theorem}

\Cref{th: final_bound_DAdagrad} nearly recovers the convergence rate of AdaGrad for static regret at the cost of an extra path length and $\mathcal{O}(\log(T)$) factor. Note that, as in \Cref{th: final_bound_DONS}, the evolutive iteration number $K_t=t$ can be chosen instead of $K=T$ to make the procedure valid for any horizon $T$ (not necessarily fixed in advance) at the cost of an extra log factor.
\\
Furthermore, \Cref{th: final_bound_DAdagrad} goes beyond the scope of \citet{zhang2017improved,zhao2021improved}, as they do not consider AdaGrad. Note that our approach is not the first to propose a dynamic regret bound for AdaGrad (see the recent work of \citealp{nazari2022dynamic}) however, our approach is, to our knowledge, the first to propose bounds on the D-C-Risk (\Cref{th: av-OT-Adagrad}), informing us on the generalization ability of OT-Adagrad.

\begin{theorem}
  \label{th: av-OT-Adagrad}
    We assume that our predictors $\hat{\xbf}$ are obtained using OT-Adagrad (\Cref{alg: dynamic_adagrad} ) with with $\varepsilon= \frac{2}{D^2}, \eta= \frac{D}{\sqrt{2}} $.
    We also assume our experts $\nu$ to be the output of  \textsc{Construct} (\Cref{alg: additional_knowledge_OMGD}) used at time $t$ with steps $\mathbf{\eta'}=(\frac{1}{\lambda j})_{j=1..K}$ and $K=T$.
    Then, dynamic cumulative risk satisfies with probability $1-2\delta$, for any $T \geq 1$, for any sequence $(\xbf_t)_{t=1..T}$ such that $\xbf_t$ is $\mathcal{F}_{t-1}$-measurable:
    \begin{align*}
        \sum_{t=1}^T L_t(\hat{\xbf}_t) - \sum_{t=1}^T L_t(\xbf_t)
        & \leq  GP_T(\nu) + \mathcal{O}\left(\sqrt{\min_{H \in \mathcal{H}} \sum_{t}\left\|\nabla_{t}\right\|_{H}^{* 2}}+ \log \frac{T}{\delta}\right).
    \end{align*}
    Note that this result still holds for any additional knowlege $\nu$ such that for any $t$, $\ell_t(\nu_{t}) - \ell_t(\xbf_t^*) = \mathcal{O}(1/t)$.
\end{theorem}

\section{Proof of OT-OMD of \Cref{sec: nonconvex}}
\label{sec: proof-nonconvex}
In this section, we will use the notation concerning Bregman divergence defined in \Cref{sec: technical_b}.

We now prove \Cref{th: OT-OMD}.

\begin{proof}
  In what follows, we write for any $t\in\{1\cdots T\}$, $\mathbb{E}_{u_t}$ the expectation under $u_t\sim \text{Unif}([0,1])$ and $\mathbb{E}_{u_{1:T}}$ the expectation under the joint distribution $(u_1, \ldots, u_T)\sim \text{Unif}([0,1])^{\otimes T}$. 
   First, we exploit that $\nu_{t} =\mathbf{0}(\ell_t)$ to get:

   \begin{align*}
    \sum_{t=1}^{T}\ell_t (\hat{\xbf}_{t}) - \inf_{\xbf\in\mathcal{K}}\ell_t(\xbf)  & = \sum_{t=1}^{T}\ell_t (\hat{\xbf}_{t}) - \ell_t (\nu_{t}) + \sum_{t=1}^{T}\ell_t (\nu_{t}) - \inf_{\xbf\in\mathcal{K}}\ell_t(\xbf) \\
    & \leq \sum_{t=1}^{T}\ell_t (\hat{\xbf}_{t}) - \ell_t (\nu_{t}) + \xi T.
   \end{align*}
   
   To deal with the remaining sum, we have, by the fundamental theorem of calculus: for any $t$,

    \begin{align*}
        \ell_t(\hat{\xbf}_{t}) - \ell_t(\nu_{t}) & = \int_{u=0}^{1} \LA \nabla\ell_t(\nu _t + u(\hat{\xbf}_{t} - \nu _t)),\hat{\xbf}_{t} - \nu _t\RA du \\
        & = \mathbb{E}_{u_t}\LB \LA \nabla_t, \hat{\xbf}_{t} - \nu_{t} \RA \RB
        \intertext{Then, summing over $T$ and integrating over $u_1,\cdots,u_T$ yields:} 
        \mathbb{E}_{u_1,\cdots,u_T}\LB \sum_{t=1}^{T}  \ell_t(\hat{\xbf}_t) - \ell_t(\nu_{t})\RB & = \mathbb{E}_{u_1,\cdots,u_T}\LB \sum_{t=1}^{T}  \LA \nabla_t, \hat{\xbf}_{t} - \nu_{t} \RA\RB.
    \end{align*}

    Then notice that for any $t$, by definition of the OMD algorithm, $\nabla R(\y_{t+1}) =\nabla R(\x_{t}) - \eta \nabla_t$, thus: 

    \begin{align*} 
      \LA \nabla_t, \hat{\xbf}_{t} - \nu_{t} \RA & = \frac{1}{\eta} \LA \nabla R(\y_{t+1})- \nabla R(\hat{\xbf}_t), \nu_{t}- \hat{\xbf}_t \RA\\
      & = \frac{1}{\eta} \LB B_R\LP\nu_{t}\mid \hat{\xbf}_t\RP - B_R\LP\nu_{t}\mid \y_{t+1}\RP + B_R\LP\hat{\xbf}_t\mid \y_{t+1}\RP \RB\\
      & \leq \frac{1}{\eta} \LB B_R\LP\nu_{t}\mid \hat{\xbf}_t\RP - B_R\LP\nu_{t}\mid \hat{\xbf}_{t+1}\RP + B_R\LP\hat{\xbf}_t\mid \y_{t+1}\RP \RB,
    \end{align*}
    the last line holding from the generalised Pythagorean theorem, as $\hat{\xbf}_{t+1}$ is the projection over $\mathcal{K}$ of $\y_{t+1}$. Adding and substracting $ B_R\LP\nu_{t-1}\mid \hat{\xbf}_{t}\RP$ and summing over $T$ then gives: 
\begin{align*} 
      \sum_{t=1}^{T}  \ell_t(\hat{\xbf}_t) - \ell_t(\nu_{t}) 
      & \leq \frac{ B_R(\nu_{0},\hat{\x}_1)}{\eta}  + \frac{1}{\eta} \sum_{t=1}^{T}\LB B_R\LP\nu_{t}\mid \hat{\xbf}_{t}\RP - B_R\LP\nu_{t-1}\mid \hat{\xbf}_{t}\RP + B_R\LP\hat{\xbf}_t\mid \y_{t+1}\RP \RB,
    \end{align*}
     Recall that $\nu_{0}= \hat{\x}_1$, then $B_R(\nu_{0},\hat{\x}_1) = 0$.
    Then, by the definition of Bregman divergence, for any $t$, 

  \begin{align*}
        B_R\LP\nu_{t}\mid \hat{\xbf}_{t}\RP - B_R\LP\nu_{t-1}\mid \hat{\xbf}_{t}\RP & = R(\nu_{t}) - R(\nu_{t-1}) - \LA \nabla R (\hat{\xbf}_{t}), \nu_{t}-\nu_{t-1} \RA\\
        & \leq R(\nu_{t}) - R(\nu_{t-1}) + \|\nabla R(\hat{\x}_t)\|\|\nu_{t}-\nu_{t-1}\|,
  \end{align*}.
  Then summing over $T$  and integrating over $u_1\cdots u_T$ yields: 

   \begin{align*}
       \mathbb{E}_{u_1\cdots u_T}\LB\sum_{t=1}^T B_R\LP\nu_{t}\mid \hat{\xbf}_{t}\RP - B_R\LP\nu_{t-1}\mid \hat{\xbf}_{t}\RP \RB & = R(\nu_{T}) - R(\nu_{0}) + \mathbb{E}_{u_1\cdots u_T}\LB  \|\nabla R(\hat{\x}_t)\|\|\nu_{t}-\nu_{t-1}\| \RB \\
        & \leq D_R^2 + 2G^*P_T. 
  \end{align*}
  where $G^*= \max_{t\in\{1\cdots T\}} \mathbb{E}_{u_1\cdots u_T} \LB\|\nabla R(\hat{\x}_t)\| \RB$.

  Finally, for any $t$, define $\z_t$ such that $B_R\LP \y_{t+1} \mid \hat{\xbf}_t\RP = \frac{1}{2}\|\y_{t+1}-\hat{\xbf}_t \|_{\z_t}^2$. Then, by definition of Bregman divergence and generalised Cauchy-Schwarz inequality:  

  \begin{align*}
    B_R\LP\hat{\xbf}_t\mid \y_{t+1}\RP + B_R\LP \y_{t+1} \mid \hat{\xbf}_t\RP &= \LA \nabla R(\hat{\xbf}_t) - \nabla R(\y_{t+1}), \hat{\xbf}_t- \y_{t+1} \RA \\
    & = \eta \LA \nabla_t, \hat{\xbf}_t- \y_{t+1}\RA  \\
    & \leq \eta \|\nabla_t\|_{\z_t}^* \|\hat{\xbf}_t- \y_{t+1}\|_{\z_t}\\
    & \leq \frac{1}{2}\LP \eta^2(\|\nabla_t\|_{\z_t}^*)^2 + \|\hat{\xbf}_t- \y_{t+1}\|_{\z_t}^2\RP
    \intertext{Then, by definition of $\z_t$, $B_R\LP \y_{t+1} \mid \hat{\xbf}_t\RP = \frac{1}{2}\|\y_{t+1}-\hat{\xbf}_t \|_{\z_t}^2$. Then,}
      B_R\LP\hat{\xbf}_t\mid \y_{t+1}\RP & \leq \frac{ \eta^2}{2}(\|\nabla_t\|_{\z_t}^*)^2 \\
      & \leq \frac{\eta^2}{2}G_R^2.
  \end{align*}
  The last line holding as we assumed the dual norms bounded by $G_R$.

  Combining those three inequalities finally yields: 

  \begin{align*}
    \mathbb{E}_{u_1,\cdots,u_T}\LB \sum_{t=1}^{T}  \ell_t(\hat{\xbf}_t) - \ell_t(\nu_{t})\RB 
    & \leq \frac{\beta^2 D^2 + 2G^*P_T(\nu)}{2\eta} + \frac{\eta}{2} G_R^2 T.
  \end{align*}
 \end{proof}

\section{Proofs of deterministic results of \Cref{sec: convex,sec: additional-OT-algos}}

\label{sec: proof_determin_results}

\subsection{Proof of \Cref{l: variation_trick}}

\begin{proof}[Proof of \cref{l: variation_trick}]
First, if $\mathrm{Perf}(t+1,H,\hat{\xbf}_{temp},\nu)<0$, then $\hat{\xbf}_{t+1}= \Pi_{\mathcal{K},H}(2m_t-\hat{\xbf}_{temp,t+1})$ and one has:

 $$
 \begin{aligned}
  \left\|\hat{\xbf}_{t+1}-\nu_{t}\right\|_H^{2} & =\left\| \Pi_{\mathcal{K},H}(2m_t-\hat{\xbf}_{temp,t+1})-\nu_{t}\right\|_H^{2} & \leq \left\| 2m_t-\hat{\xbf}_{temp,t+1}-\nu_{t}\right\|_H^2  \\
  & =\|\hat{\xbf}_{temp,t+1} - \nu_{t-1}\|_H^2.
  \end{aligned}
 $$

   The last line holding thanks to the definition of $m_t$. Second, if $\mathrm{Perf}(t,H,\hat{\xbf}_{temp},\nu)\geq0$, we use:

 \begin{lemma}
 \label{l: perf_lemma}
  We have
 $\forall t \geqslant 0,\left\|\hat{\xbf}_{temp,t+1}-\nu_{t}\right\|_H^{2}=\left\|\hat{\xbf}_{temp,t+1}-\nu_{t-1}\right\|_H^{2}-2 \mathrm{Perf}(t,H, \hat{\xbf}_{temp}, \nu).$

 \end{lemma}

 \begin{proof}[Proof of \cref{l: perf_lemma}]
  Recall that $m_{t}=\frac{\nu_{t}+\nu_{t-1}}{2}$. We have:
\begin{align*}
  \left\|\hat{\xbf}_{temp,t+1}-\nu_{t}\right\|_H^{2} & =\left\| \hat{\xbf}_{temp,t+1}-m_{t}+m_{t}-\nu_{t}\right\|_H^{2} \\ & =\left\| \hat{\xbf}_{temp,t+1}-m_{t} \right\|_H^{2}-\mathrm{Perf}(t,H, \hat{\xbf}_{temp}, \nu)+\frac{\| \nu_{t-1}-\nu_{t}\|_H^2}{4}.
\end{align*}

And $\left\|\hat{\xbf}_{temp,t+1}-\nu_{t-1} \right\|_H^2=\left\| \hat{\xbf}_{temp,t+1}-m_{t} \right\|_H^{2}+\mathrm{Perf}(t,H, \hat{\xbf}_{temp}, \nu)+\frac{\left\|\nu_{t}-\nu_{t-1}\right\|_H^{2}}{4}.$

Thus,
 $
 \left\|\hat{\xbf}_{temp,t+1}-\nu_{t}\right\|_H^{2}=\left\|\hat{\xbf}_{temp,t+1}-\nu_{t-1}\right\|_H^{2}-2 \mathrm{Perf}(t,H, \hat{\xbf}_{temp}, \nu).
 $
 \end{proof}
Finally,
 \begin{align*}
  \left\|\hat{\xbf}_{t+1}-\nu_{t}\right\|_H^{2} & =\left\| \Pi_{\mathcal{K},H}(\hat{\xbf}_{temp,t+1})-\nu_{t}\right\|_H^{2} \leq \left\| \hat{\xbf}_{temp,t+1}-\nu_{t}\right\|_H^2  \\
   & =\|\hat{\xbf}_{temp,t+1} - \nu_{t-1}\|_H^2 - 2\mathrm{Perf}(t,H,\hat{\xbf}_{temp},\nu) & \text{(by \cref{l: perf_lemma})}\\
  & \leq \|\hat{\xbf}_{temp,t+1} - \nu_{t-1}\|_H^2 .
  \end{align*}

   The last line holding because our performance is positive in this case. This concludes the proof.
 \end{proof}

\subsection{A general route of proof}

We exhibit in \cref{eq: general_scheme} a general pattern of proof we use several times in what follows to bound the dynamic regret. 

\begin{align}
  \label{eq: general_scheme}
  \sum_{t=1}^T \ell_t(\hat{\xbf}_t) - \sum_{t=1}^T \inf_{\xbf \in \mathcal{K}}\ell_t(\xbf) = \underbrace{\sum_{t=1}^T \ell_t(\hat{\xbf}_t) - \sum_{t=1}^T \ell_t(\nu_{t-1})}_{= (A)}
  + \underbrace{\sum_{t=1}^T \ell_t(\nu_{t-1}) - \sum_{t=1}^T \ell_t(\nu_{t})}_{=(B)} + \underbrace{\sum_{t=1}^T \ell_t(\nu_{t}) - \sum_{t=1}^T \ell_t(\xbf_t^*)}_{=(C)} .
\end{align}

Those terms are dealt as follows.

\begin{itemize}
  \item (A) is controlled by the effect of \textsc{Adjust} on OGD,ONS,Adagrad. It allows to transform the static guarantees of those algorithms (as stated in \citealp{hazan2019introduction}) into dynamic ones.
  \item (B) is controlled by the convexity assumptions made on the $\ell_t$s and involve terms like $P_T,S_T$.
  \item (C) is handled by the way we designed $\nu$.
\end{itemize}

Our proofs in the rest of this section are based on this general scheme.

  Note that we used the sequence $\xbf^*= (\xbf_t^*)_{t\geq 1}$ as comparators here in order to control (C) via the \textsc{Construct} algorithm. This has two implications: \textit{(i)} our results then holds when using any comparator sequence as we control the worst case dynamic regret and \textit{(ii)} we can also involve directly any other comparator sequence $\xbf$ within the proof, at the cost of letting (C) unconstrained. We would then need an algorithm different from \textsc{Construct} in order to make $\nu_{t}$ close to $\xbf_t$.

\begin{remark}
  In the rest of this section, we use the shortcut $\nabla_t:= \nabla \ell_t(\hat{\xbf}_t)$ and refer to $\xbf_t^*$ as an element of $\operatorname{argmin}_{\xbf \in \mathcal{K}}\ell_t(\xbf)$.
\end{remark}

\subsection{Proof of Thm \ref{th: D_regret_bound_general}}

\begin{proposition}
    \label{prop: D-regret_convex}
    \noindent  The sequence of predictors $(\hat{\xbf}_t)_{t\geq0} $ obtained through DOGD on a closed convex $\mathcal{K}$ with steps $\mathbf{\eta}= (\eta_t)_{t=1..T}$ with regards to the experts $\mathbf{\nu}=(\nu_{t})_{t=0..T}\in\mathcal{K}^T$ satisfies :

        \begin{align*}
            \sum_{t=1}^T \ell_t(\hat{\xbf}_t) - \sum_{t=1}^T \ell_t(\nu_{t-1})
            & \leq \frac{D^2}{2\eta_T} + \frac{G^2}{2}\sum_{t=1}^T\eta_t.
        \end{align*}
\end{proposition}

\begin{proof}

    \noindent We fix $t\geq 0$. For the sake of clarity, we rename $\hat{\xbf}_{temp}:= \hat{\xbf}_{temp,t+1}= \hat{\xbf}_{t} - \eta_t \nabla \ell_t(\hat{\xbf}_t) $ (where $\hat{\xbf}_{temp,t+1}$ is defined in \cref{alg: dynamic_OGD}).

    \noindent Thanks to convexity of the losses, one has:

    \begin{align*}
            \sum_{t=1}^T \ell_t(\hat{\xbf}_t) - \sum_{t=1}^T \ell_t(\nu_{t-1}) & \leq \sum_{t=1}^T \langle\nabla\ell_t(\hat{\xbf}_t), \hat{\xbf}_t- \nu_{t-1}\rangle.
     \end{align*}

      \noindent To control this last sum, our intermediary goal is now to control $||\hat{\xbf}_{t+1} - \nu_{t}||^2$ in function of $\|\hat{\xbf}_t- \nu_{t-1}\|^2$.
      To do so, we first exploit \cref{l: variation_trick} which stipulates that $ ||\hat{\xbf}_{t+1} - \nu_{t}||^2 \leq \|\hat{\xbf}_{temp} - \nu_{t-1}\|^2 $.
      Then we control $\left\langle\nabla \ell_{t}\left(\hat{\xbf}_{t}\right), \hat{\xbf}_{t}-\nu_{t-1}\right\rangle $.

    \noindent One has:

    $$
    \begin{aligned}
      ||\hat{\xbf}_{t+1} - \nu_{t}||^2 & \leq \|\hat{\xbf}_{temp} - \nu_{t-1}\|^2\\
      &  = \left\|\hat{\xbf}_{t}-\eta_{t} \nabla \ell_{t}\left(\hat{\xbf}_{t}\right)-\nu_{t-1}\right\|^{2} \\
    &=\left\|\hat{\xbf}_{t}-\nu_{t-1}\right\|^{2}-2 \eta_{t}\left\langle\nabla \ell_{t}\left(\hat{\xbf}_{t}\right), \hat{\xbf}_{t}-\nu_{t-1}\right\rangle+\eta_{t}^{2}\left\|\nabla \ell_{t}\left(\hat{\xbf}_{t}\right)\right\|^{2}
    \end{aligned}
    $$

    \noindent Hence: $$\left\|\hat{\xbf}_{t+1}-\nu_{t} \right\|^{2} \leqslant\left\| \hat{\xbf}_{t}-\nu_{t-1} \right\|^{2}-2 \eta_{t}\left\langle\nabla \ell_{t}\left(\hat{\xbf}_{t}\right), \hat{\xbf}_{t}-\nu_{t-1}\right\rangle+\eta_{t}^{2} G^{2}.$$

    \noindent So:

    $$\left\langle\nabla \ell_{t}\left(\hat{\xbf}_{t}\right), \hat{\xbf}_{t}-\nu_{t-1}\right\rangle \leqslant \frac{\left\|\hat{\xbf}_{t}-\nu_{t-1}\right\|^{2}-\left\|\hat{\xbf}_{t+1}- \nu_{t}\right\|^{2}}{2 \eta_{t}}+\frac{\eta_{t} G^{2}}{2}.$$
    \noindent Summing on $t$, gives

    \begin{align*}
      \sum_{t=1}^T \ell_t(\hat{\xbf}_t) - \sum_{t=1}^T \ell_t(\nu_{t-1}) & \leq \sum_{t=1}^T \left\|\hat{\xbf}_{t}-\nu_{t-1}\right\|^{2}\left( \frac{1}{2\eta_{t}} - \frac{1}{2\eta_{t-1}}  \right) + \frac{\eta_{t} G^{2}}{2} \\
       \leq \frac{D^{2}}{2 \eta_{T}} + \frac{G^2}{2}\sum_{t=1}^T\eta_t.
    \end{align*}
    Hence the final result.

\end{proof}

Now we are able to prove our result:

\paragraph{Proof of \cref{th: D_regret_bound_general}}

\begin{proof}

We control the terms presented in \cref{eq: general_scheme}. \cref{prop: D-regret_convex} ensures us that:

\begin{align*}
       (A)
       & \leq  \frac{D^2}{2\eta_T} + \frac{G^2}{2}\sum_{t=1}^T\eta_t \\
       & \leq \frac{3}{2}GD\sqrt{T},
        \end{align*}

        \noindent The last line holding thanks to the definition of $\eta$ and that $\sum_{t=1}^T \frac{1}{\sqrt{t}} \leq 2\sqrt{T}$.

    \noindent We now have to deal with (B) and (C) of \cref{eq: general_scheme}.

    \noindent (B) is handled using the bounded gradient property of $\ell_t$ for any $t:$

    \begin{align*}
         \ell_t(\nu_{t-1}) -  \ell_t(\nu_{t}) & \leq \nabla\ell_t(\nu_{t-1})^\top (\nu_{t-1}-\nu_{t}) - \lambda ||\nu_{t}-\nu_{t-1}||^2 \\
         & \leq G|| \nu_{t}-\nu_{t-1}||.
         \intertext{Summing over all $t$ gives us :}
         (B) & \leq GP_T(\nu).
    \end{align*}

    \noindent To deal with (C), we exploit \cref{l: GD_add_know}.  Indeed, our choice of steps ensure us that, at each time $t$:
    \begin{align*}
      \ell_t(\nu_{t}) - \ell_t(\xbf_t^*) & \leq \frac{3}{2\sqrt{K_t}}GD.
    \end{align*}

    \noindent Finally, summing over $t$ and recalling that $K_t=t$ yields:

    \begin{align*}
        (C) & \leq \frac{3}{2}GD \sum_{t=1}^T \frac{1}{\sqrt{t}}\\
        &\leq 3GD \sqrt{T}.
    \end{align*}

    \noindent Combining the bounds of (A),(B),(C) concludes the proof.

    \end{proof}

\subsection{Proof of Thm \ref{th: final_bound_DONS}}

\noindent  We need first to introduce on exp-concave funtion.

\begin{definition}
  A function $f:\mathbb{R}^n\rightarrow \mathbb{R}$ is $\alpha$ exp-concave over a convex $\mathcal{K}$ if the function $g= \exp(-\alpha f)$ is concave on $\mathcal{K}$.
\end{definition}

\noindent One also recalls the following lemma coming from \cite[Lemma 4.3]{hazan2019introduction}

\begin{lemma}
  \label{l: exp_concave_property}
  Let $f: \mathcal{K} \rightarrow \mathbb{R}$ be an $\alpha$-exp-concave function, and $D, G$ denote the diameter of $\mathcal{K}$ and a bound on the (sub)gradients of $f$ respectively. The following holds for all $\gamma \leq \frac{1}{2} \min \left\{\frac{1}{4 G D}, \alpha\right\}$ and all $\xbf, \ybf \in \mathcal{K}$ :
$$
f(\xbf) \geq f(\ybf)+\nabla f(\ybf)^{\top}(\xbf-\ybf)+\frac{\gamma}{2}(\xbf-\ybf)^{\top} \nabla f(\ybf) \nabla f(\ybf)^{\top}(\xbf-\ybf) .
$$
\end{lemma}

One now states a key preliminary result of this section (\cref{prop: dynamic_ONS}) whoch exploits the exp-concavity property.

\begin{proposition}
  \label{prop: dynamic_ONS}
  We assume our loss functions $\ell_t$ to be $\alpha$ exp-concave.
  Let $\{\hat{\xbf}_t\}$ being the output of D-ONS (\cref{alg: dynamic_ONS}) with $\gamma= \frac{1}{2} \min \left\{\frac{1}{ G D}, \alpha \right\}, \; \varepsilon= \frac{1}{\gamma^2 D^2}  $. We then have, for $T>4$ and any additional knowledge $\nu$:

  \[ \sum_{t=1}^T \ell_t(\hat{\xbf}_t) - \ell_t(\nu_{t-1}) \leq 2\left( \frac{1}{\alpha} + GD    \right) d \log(T).  \]

\end{proposition}

\begin{proof}
  The proof is similar to the one of \citep[Thm 4.5]{hazan2019introduction} which holds for static regret. We prove \cref{l: dynamic_ONS} which is an adaptation of \citep[Lemma 4.6]{hazan2019introduction}.

  \begin{lemma}
    \label{l: dynamic_ONS}
    Let $\{\hat{\xbf}_t\}$ being the output of \cref{alg: dynamic_ONS} with $\gamma= \frac{1}{2} \min \left\{\frac{1}{ G D},\alpha \right\}, \; \varepsilon= \frac{1}{\gamma^2 D^2}  $. We then have, for $T>4$ and any additional knowledge $\nu$:

    \[ \sum_{t=1}^T \ell_t(\hat{\xbf}_t) - \ell_t(\nu_{t-1}) \leq \left( \frac{1}{\alpha} + GD    \right)\left( 1 + \sum_{t=1}^T \nabla_{t} A_t^{-1} \nabla_{t}^{\top}    \right).   \]

  \end{lemma}

  \begin{proof}
    We fix $t\geq 1$ and we first apply \cref{l: exp_concave_property}:

    \[ \ell_t(\hat{\xbf}_t) - \ell_t(\nu_{t-1}) \leq \nabla_t^\top (\hat{\xbf}_t- \nu_{t-1}) - \frac{\gamma}{2} (\hat{\xbf}_t- \nu_{t-1})^\top \nabla_t \nabla_t^\top (\hat{\xbf}_t- \nu_{t-1})   \]

    \noindent Recalling the definition of $\hat{\xbf}_{temp,t+1}$, substracting by $\nu_{t-1}$ and multiplying by $A_t$ gives us:

\begin{align}
  \label{eq: ONS_hazan_4_1}
  \hat{\xbf}_{temp,t+1}- \nu_{t-1}& = \hat{\xbf}_{t} - \nu_{t-1} - \frac{1}{\gamma} A_{t}^{-1} \nabla_{t}\\
  \intertext{and: }
  \label{eq: ONS_hazan_4_2}
  A_{t}\left(\hat{\xbf}_{temp,t+1}-\nu_{t-1}\right) & =A_{t}\left(\hat{\xbf}_{t}-\nu_{t-1}\right)-\frac{1}{\gamma} \nabla_{t}
\end{align}

Multiplying the transpose of \cref{eq: ONS_hazan_4_1} by \cref{eq: ONS_hazan_4_2} gives us:
\begin{align}
  \label{eq: ONS_hazan_4_3}
  \left(\hat{\xbf}_{temp,t+1}-\nu_{t-1}\right)^{\top} A_{t}\left(\hat{\xbf}_{temp,t+1}-\nu_{t-1}\right)=
  \left(\hat{\xbf}_{t}-\nu_{t-1}\right)^{\top} A_{t}\left(\hat{\xbf}_{t}-\nu_{t-1}\right)-\frac{2}{\gamma} \nabla_{t}^{\top}\left(\hat{\xbf}_{t}-\nu_{t-1}\right)+\frac{1}{\gamma^{2}} \nabla_{t}^{\top} A_{t}^{-1} \nabla_{t} .
\end{align}

\noindent Our goal is to lower bound the term on left hand-side of this equality. To do so, we first remark

$$
\begin{gathered}
\left(\hat{\xbf}_{temp,t+1}-\nu_{t-1}\right)^{\top} A_{t}\left(\hat{\xbf}_{temp,t+1}-\nu_{t-1}\right) \quad=\left\|\hat{\xbf}_{temp,t+1}-\nu_{t-1}\right\|_{A_{t}}^{2}
\end{gathered}
$$
\noindent Because $A_t$ is a positive definite matrix, \cref{l: variation_trick} holds, which allows us to say that  $\left\|\hat{\xbf}_{temp,t+1}-\nu_{t-1}\right\|_{A_{t}}^{2} \geq \left\|\hat{\xbf}_{t+1}-\nu_{t}\right\|_{A_{t}}^{2}$. Thus:

\begin{align*}
  \left(\hat{\xbf}_{temp,t+1}-\nu_{t-1}\right)^{\top} A_{t}\left(\hat{\xbf}_{temp,t+1}-\nu_{t-1}\right) &\geq\left\|\hat{\xbf}_{t+1}-\nu_{t}\right\|_{A_{t}}^{2} \\
    & =\left(\hat{\xbf}_{t+1}-\nu_{t}\right)^{\top} A_{t}\left(\hat{\xbf}_{t+1}-\nu_{t}\right)
\end{align*}

\noindent This fact together with \cref{eq: ONS_hazan_4_3} gives:
$$
\begin{aligned}
\nabla_{t}^{\top}\left(\hat{\xbf}_{t}-\nu_{t-1}\right) & \leq \frac{1}{2 \gamma} \nabla_{t}^{\top} A_{t}^{-1} \nabla_{t}+\frac{\gamma}{2}\left(\hat{\xbf}_{t}-\nu_{t-1}\right)^{\top} A_{t}\left(\hat{\xbf}_{t}-\nu_{t-1}\right) \\
&-\frac{\gamma}{2}\left(\hat{\xbf}_{t+1}-\nu_{t}\right)^{\top} A_{t}\left(\hat{\xbf}_{t+1}-\nu_{t}\right) .
\end{aligned}
$$
\noindent Now, summing up over $t=1$ to $T$ we get that
$$
\begin{aligned}
&\sum_{t=1}^{T} \nabla_{t}^{\top}\left(\hat{\xbf}_{t}-\nu_{t-1}\right) \leq \frac{1}{2 \gamma} \sum_{t=1}^{T} \nabla_{t}^{\top} A_{t}^{-1} \nabla_{t}+\frac{\gamma}{2}\left(\xbf_{1}-\nu_{0}\right)^{\top} A_{1}\left(\xbf_{1}-\nu_{0}\right) \\
&\quad+\frac{\gamma}{2} \sum_{t=2}^{T}\left(\hat{\xbf}_{t}-\nu_{t-1}\right)^{\top}\left(A_{t}-A_{t-1}\right)\left(\hat{\xbf}_{t}-\nu_{t-1}\right) \\
&\quad-\frac{\gamma}{2}\left(\hat{\xbf}_{T+1}-\nu_{t}\right)^{\top} A_{T}\left(\hat{\xbf}_{T+1}-\nu_{t}\right) \\
&\leq \frac{1}{2 \gamma} \sum_{t=1}^{T} \nabla_{t}^{\top} A_{t}^{-1} \nabla_{t}+\frac{\gamma}{2} \sum_{t=1}^{T}\left(\hat{\xbf}_{t}-\nu_{t-1}\right)^{\top} \nabla_{t} \nabla_{t}^{\top}\left(\hat{\xbf}_{t}-\nu_{t-1}\right) \\
&+\frac{\gamma}{2}\left(\xbf_{1}-\nu_{0}\right)^{\top}\left(A_{1}-\nabla_{1} \nabla_{1}^{\top}\right)\left(\xbf_{1}-\nu_{0}\right)
\end{aligned}
$$
In the last inequality we use the fact that $A_{t}-A_{t-1}=\nabla_{t} \nabla_{t}^{\top}$, and the fact that the matrix $A_{T}$ is PSD to bound the last term before the inequality by 0. Thus,
$$
\sum_{t=1}^{T} \ell_{t}(\hat{\xbf}_t)- \ell_t(\nu_{t-1}) \leq \frac{1}{2 \gamma} \sum_{t=1}^{T} \nabla_{t}^{\top} A_{t}^{-1} \nabla_{t}+\frac{\gamma}{2}\left(\xbf_{1}-\nu_{0}\right)^{\top}\left(A_{1}-\nabla_{1} \nabla_{1}^{\top}\right)\left(\xbf_{1}-\nu_{0}\right)
$$

\noindent Using that $ A_{1}-\nabla_{1} \nabla_{1}^{\top}= \varepsilon I_n, \; \varepsilon= \frac{1}{\gamma^2 D^2}$  and that $\mathcal{K}$ has a finite diameter $D$ gives us :

$$
\begin{aligned}
\sum_{t=1}^{T} \ell_{t}(\hat{\xbf}_t)- \ell_t(\nu_{t-1}) & \leq \frac{1}{2 \gamma} \sum_{t=1}^{T} \nabla_{t}^{\top} A_{t}^{-1} \nabla_{t}+\frac{\gamma}{2} D^{2} \varepsilon \\
& \leq \frac{1}{2 \gamma} \sum_{t=1}^{T} \nabla_{t}^{\top} A_{t}^{-1} \nabla_{t}+\frac{1}{2 \gamma}
\end{aligned}
$$
Since $\gamma=\frac{1}{2} \min \left\{\frac{1}{ G D}, \alpha\right\}$, we have $\frac{1}{\gamma} \leq 2\left(\frac{1}{\alpha}+G D\right)$. This gives the lemma.

  \end{proof}

\noindent The rest of the proof now follows the exact same route than \cite[Thm 4.5]{hazan2019introduction}.

\textbf{Proof of \cref{prop: dynamic_ONS}}
 First we show that the term $\sum_{t=1}^{T} \nabla_{t}^{\top} A_{t}^{-1} \nabla_{t}$ is upper bounded by a telescoping sum. Notice that
$$
\nabla_{t}^{\top} A_{t}^{-1} \nabla_{t}=A_{t}^{-1} \bullet \nabla_{t} \nabla_{t}^{\top}=A_{t}^{-1} \bullet\left(A_{t}-A_{t-1}\right)
$$
where for matrices $A, B \in \mathbb{R}^{n \times n}$ we denote by $A \bullet B=\sum_{i=1}^{n} \sum_{j=1}^{n} A_{i j} B_{i j}=$ $\operatorname{Tr}\left(A B^{\top}\right)$, which is equivalent to the inner product of these matrices as vectors in $\mathbb{R}^{n^{2}}$.

For real numbers $a, b \in \mathbb{R}_{+}$, the first order Taylor expansion of the logarithm of $b$ at $a$ implies $a^{-1}(a-b) \leq \log \frac{a}{b}$. An analogous fact holds for positive semidefinite matrices, i.e., $A^{-1} \bullet(A-B) \leq \log \frac{|A|}{|B|}$, where $|A|$ denotes the determinant of the matrix $A$ (this is proved in \citealp[Lemma 4.7]{hazan2019introduction}). Using this fact we have
$$
\begin{aligned}
\sum_{t=1}^{T} \nabla_{t}^{\top} A_{t}^{-1} \nabla_{t} &=\sum_{t=1}^{T} A_{t}^{-1} \bullet \nabla_{t} \nabla_{t}^{\top} \\
&=\sum_{t=1}^{T} A_{t}^{-1} \bullet\left(A_{t}-A_{t-1}\right) \\
& \leq \sum_{t=1}^{T} \log \frac{\left|A_{t}\right|}{\left|A_{t-1}\right|}=\log \frac{\left|A_{T}\right|}{\left|A_{0}\right|}
\end{aligned}
$$
Since $A_{T}=\sum_{t=1}^{T} \nabla_{t} \nabla_{t}^{\top}+\varepsilon I_{n}$ and $\left\|\nabla_{t}\right\| \leq G$, the largest eigenvalue of $A_{T}$ is at most $T G^{2}+\varepsilon$. Hence the determinant of $A_{T}$ can be bounded by $\left|A_{T}\right| \leq\left(T G^{2}+\varepsilon\right)^{d}$. Hence recalling that $\varepsilon=\frac{1}{\gamma^{2} D^{2}}$ and $\gamma=\frac{1}{2} \min \left\{\frac{1}{G D}, \alpha\right\}$, for $T>4$
$$
\sum_{t=1}^{T} \nabla_{t}^{\top} A_{t}^{-1} \nabla_{t} \leq \log \left(\frac{T G^{2}+\varepsilon}{\varepsilon}\right)^{d} \leq d \log \left(T G^{2} \gamma^{2} D^{2}+1\right) \leq d \log T
$$
Plugging into \cref{l: dynamic_ONS} we obtain
$$
\sum_{t=1}^{T} \ell_{t}(\hat{\xbf}_t)- \ell_t(\nu_{t-1}) \leq\left(\frac{1}{\alpha}+G D\right)(d \log T+1)
$$
which implies the theorem for $d>1, T \geq 4$.

\end{proof}

We now can prove \cref{th: final_bound_DONS}.

\paragraph{Proof of \cref{th: final_bound_DONS}.}

\begin{proof}

  We control the terms presented in \cref{eq: general_scheme}. To deal with (A), we exploit \cref{prop: dynamic_ONS} knowing that a $\lambda$-strongly convex function with its gradient bounded by $G$ is $\lambda/G^2$ exp-concave:

  \begin{align*}
     (A)
     & \leq  2\left( \frac{G^2}{\lambda} + GD    \right) d (1+\log(T))
      \end{align*}

    \noindent We now have to deal with (B) and (C) of \cref{eq: general_scheme}.

    \noindent (B) is handled using the strong convexity of $\ell_t$ for any $t:$

    \begin{align*}
         \ell_t(\nu_{t-1}) -  \ell_t(\nu_{t}) & \leq \nabla\ell_t(\nu_{t-1})^\top (\nu_{t-1}-\nu_{t}) - \lambda ||\nu_{t}-\nu_{t-1}||^2 & \\
         & \leq ||\nabla\ell_t(\nu_{t-1})||.|| \nu_{t}-\nu_{t-1}|| - \lambda ||\nu_{t}-\nu_{t-1}||^2 & \text{Cauchy-Schwarz}\\
         & \leq G|| \nu_{t}-\nu_{t-1}|| - \lambda ||\nu_{t}-\nu_{t-1}||^2.
         \intertext{Summing over all $t$ gives us :}
         (B) & \leq GP_T(\nu) - \lambda S_T(\nu).
    \end{align*}

    \noindent To deal with (C), we exploit \cref{l: GD_add_know}.  Indeed, our choice of steps ensure us that at each step $j$: $ \frac{1}{\eta'_j} - \lambda =\lambda\left( j -1\right) = \frac{1}{\eta_{j-1}'}  $.
    We have at each time $t$:
    \begin{align*}
      \ell_t(\nu_{t}) - \ell_t(\xbf_t^*) & \leq \frac{G^2}{K} \sum_{j=1}^K  \eta'_j = \frac{G^2}{\lambda K} \sum_{j=1}^K \frac{1}{j} \\
      & \leq \frac{G^2 (1+\log(K))}{\lambda K}.
    \end{align*}

    \noindent Finally:

    \begin{align*}
        (C) & \leq T \frac{G^2 (1+\log(K))}{\lambda K} \\
        & = \frac{G^2}{\lambda}(1+\log(T))
    \end{align*}

    \noindent The last line holding because $K = T$.

    \noindent Combining the bounds on (A),(B),(C) concludes the proof.

    \end{proof}

\subsection{Proof of Thm \ref{th: final_bound_DAdagrad}}

We first start with a key result for our study of dynamic Adagrad.

\begin{proposition}
\label{prop: dynamic_adagrad}
We assume our loss functions $\ell_t$ to be convex.
Let $\{\hat{\xbf}_t\}$ being the output of D-Adagrad (\cref{alg: dynamic_adagrad}) with $\varepsilon= \frac{2}{D^2}, \eta= \frac{D}{\sqrt{2}} $. We then have, for any additional knowledge $\nu$:
$$
\sum_{t=1}^T \ell_t(\hat{\xbf}_t)- \ell_t(\nu_{t-1}) \leq \sqrt{2} D\left( 1 + \sqrt{\min_{H \in \mathcal{H}} \sum_{t}\left\|\nabla_{t}\right\|_{H}^{* 2}} \right)
$$
where $\mathcal{H}:= \left\{ X\in \mathbb{R}^{n\times n} \mid Tr(X)\leq 1, X \succeq 0    \right\}$ and  for a fixed $H$, $||\xbf||^{* 2}_H = \xbf^T H^{-1} \xbf$ where $H^{-1}$ refers to the Moore-Penrose pseudoinverse.
\end{proposition}

\begin{proof}

The proof follows the route of \cite[Thm 5.12]{hazan2019introduction} for the full-matrix version of Adagrad. As for dynamic ONS, our only work consists in modifying a lemma of Hazan's proof (\citealp[Lemma 5.13]{hazan2019introduction}), the rest holding similarly.

For the sake of completeness, we state all the lemma of interest in this proof, most of them are directly extracted from \citep[Sec.5.6]{hazan2019introduction}. We start with \citep[Lemma 11]{hazan2019introduction}.

\begin{lemma}
  For $H_T$ the last output of Adagrad, we have

  \[   \sqrt{\min_{H \in \mathcal{H}} \sum_{t}\left\|\nabla_{t}\right\|_{H}^{* 2}} = \textbf{Tr}(H_T)  \]
\end{lemma}

We present now our lemma of interest (\citealp[Lemma 5.13]{hazan2019introduction})

\begin{lemma}
  \label{l: hazan_lemma_5_13}
  \[\sum_{t=1}^T \ell_t(\hat{\xbf}_t) - \ell_t(\nu_{t-1}) \leq 2D + \frac{\eta}{2} \left(G_{T} \bullet H_{T}^{-1}+\operatorname{Tr}(H_T)\right)+\frac{1}{2 \eta} \sum_{t=1}^{T}\left(\hat{\xbf}_{t} -\hat{\nu_{t-1}}\right)^{\top}\left(H_{t}-H_{t-1}\right)\left(\hat{\xbf}_{t}-\nu_{t-1}\right) . \]
\end{lemma}

\begin{proof}

First, recall that $ \sum_{t=1}^T \ell_t(\hat{\xbf}_t) - \ell_t(\nu_{t-1}) \leq \sum_{t=1}^{T} \nabla_{t}^{\top}\left(\hat{\xbf}_{t}-\nu_{t-1}\right)$.

\noindent By the definition of $\hat{\xbf}_{temp,t+1}$ :
\begin{align}
  \label{eq: Adagrad_hazan_5_8}
  \hat{\xbf}_{temp,t+1}-\nu_{t-1}=\hat{\xbf}_{t}-\nu_{t-1}-\eta H_{t}^{-1} \nabla_{t} \\
  \intertext{and multipying by $H_t$ gives:}
  \label{eq: Adagrad_hazan_5_9}
  H_{t}\left(\hat{\xbf}_{temp,t+1}-\nu_{t-1}\right)=H_{t}\left(\hat{\xbf}_{t}-\nu_{t-1}\right)-\eta \nabla_{t} .
\end{align}

\noindent Multiplying the transpose of \cref{eq: Adagrad_hazan_5_8} by \cref{eq: Adagrad_hazan_5_9} we get
\begin{multline}
  \label{eq: Adagrad_hazan_5_10}
  \left(\hat{\xbf}_{temp,t+1}-\nu_{t-1}\right)^{\top} H_{t}\left(\hat{\xbf}_{temp,t+1}-\nu_{t-1}\right) \\ =
  \left(\hat{\xbf}_{t}-\nu_{t-1}\right)^{\top} H_{t}\left(\hat{\xbf}_{t}-\nu_{t-1}\right)-2 \eta \nabla_{t}^{\top}\left(\hat{\xbf}_{t}-\nu_{t-1}\right)  +\eta^{2} \nabla_{t}^{\top} H_{t}^{-1} \nabla_{t} .
\end{multline}

\noindent Focusing on the left-hand side of the equality, one remarks that:
$$
\left(\hat{\xbf}_{temp,t+1}-\nu_{t-1}\right)^{\top} H_{t}\left(\hat{\xbf}_{temp,t+1}-\nu_{t-1}\right)=\left\|\hat{\xbf}_{temp,t+1}-\nu_{t-1}\right\|_{H_{t}}^{2}
$$

Since $H_t$ is a PD matrix, one can apply \cref{l: variation_trick} to obtain that $\| \hat{\xbf}_{t+1} - \nu_{t}\|_{H_t}^2 \leq \| \hat{\xbf}_{temp,t+1}- \nu_{t-1}\|_{H_t}^2$.

\noindent Applying this result gives:

\[ \left(\hat{\xbf}_{temp,t+1}-\nu_{t-1}\right)^{\top} H_{t}\left(\hat{\xbf}_{temp,t+1}-\nu_{t-1}\right)  \geq\left\|\hat{\xbf}_{t+1}-\nu_{t}\right\|_{H_{t}}^{2} \]

\noindent
This fact together with \cref{eq: Adagrad_hazan_5_10} gives
$$
\nabla_{t}^{\top}\left(\hat{\xbf}_{t}-\nu_{t-1}\right) \leq \frac{\eta}{2} \nabla_{t}^{\top} H_{t}^{-1} \nabla_{t}+\frac{1}{2 \eta}\left(\left\|\hat{\xbf}_{t}-\nu_{t-1}\right\|_{H_{t}}^{2}-\left\|\hat{\xbf}_{t+1}-\nu_{t}\right\|_{H_{t}}^{2}\right)
$$
Now, summing up over $t=1$ to $T$ we get that
\begin{multline*}
\sum_{t=1}^{T} \nabla_{t}^{\top}\left(\hat{\xbf}_{t}-\nu_{t-1}\right) \leq \\
\frac{\eta}{2} \sum_{t=1}^{T} \nabla_{t}^{\top} H_{t}^{-1} \nabla_{t}+\frac{1}{2 \eta}\left\|\xbf_{1}-\nu_{0}\right\|_{H_{0}}^{2} +\frac{1}{2 \eta} \sum_{t=1}^{T}\left(\left\|\hat{\xbf}_{t}-\nu_{t-1}\right\|_{H_{t}}^{2}-\left\|\hat{\xbf}_{t}-\nu_{t-1}\right\|_{H_{t-1}}^{2}\right)
-\frac{1}{2 \eta}\left\|\hat{\xbf}_{t+1}-\nu_{t}\right\|_{H_{T}}^{2} \\
\leq \frac{\eta}{2} \sum_{t=1}^{T} \nabla_{t}^{\top} H_{t}^{-1} \nabla_{t}+ \sqrt{2}D +\frac{1}{2 \eta} \sum_{t=1}^{T}\left(\hat{\xbf}_{t}-\nu_{t-1}\right)^{\top}\left(H_{t}-H_{t-1}\right)\left(\hat{\xbf}_{t}-\nu_{t-1}\right)  .
\end{multline*}
In the last inequality we used the fact that $\varepsilon=\frac{2}{D^2}$ and bounded $\|\ \xbf_1 - \nu_{0}  \|$ by $D^2$ .

We now prove that $\sum_{t=1}^{T} \nabla_{t}^{\top} H_{t}^{-1} \nabla_{t} \leq \left(G_{T} \bullet H_{T}^{-1}+\operatorname{Tr}(H_T)\right)$. To this end, define the functions
$$
\Psi_{t}(H)=\nabla_{t} \nabla_{t}^{\top} \bullet H^{-1}, \Psi_{0}(H)=\operatorname{Tr}(H) .
$$
By definition, $H_{t}$ is the minimizer of $\sum_{i=0}^{t} \Psi_{i}$ over $\mathcal{H}$ which can be related to a FTL strategy. Thus, using \citep[Lemma 5.4]{hazan2019introduction}, we have that
$$
\begin{aligned}
\sum_{t=1}^{T} \nabla_{t}^{\top} H_{t}^{-1} \nabla_{t} &=\sum_{t=1}^{T} \Psi_{t}\left(H_{t}\right) \\
\leq  &\sum_{t=1}^{T} \Psi_{t}\left(H_{T}\right)+\Psi_{0}\left(H_{T}\right)-\Psi_{0}\left(H_{0}\right) \\
&=G_{T} \bullet H_{T}^{-1}+\operatorname{Tr}\left(H_{T}\right)
\end{aligned}
$$

This concludes the proof
\end{proof}

\noindent \cref{l: hazan_lemma_5_13} gives us two terms to be bounded. To do so, we use \cite[Lemmas 5.14,5.15]{hazan2019introduction} to conclude the proof. Those lemmas are gathered below.

\begin{lemma}
  For \cref{alg: dynamic_adagrad}, the following holds
  $$
  G_{T} \bullet H_{T}^{-1} \leq \operatorname{Tr}\left(H_{T}\right) .
  $$
\end{lemma}

\begin{lemma}
  Recall that $D$ the Euclidean diameter of $\mathcal{K}$. Then the following bound holds,
   $\quad \sum_{t=1}^{T}\left\|\xbf_{t}-\xbf^{\star}\right\|_{H_{t}-H_{t-1}}^{2} \leq D^{2} \operatorname{Tr}\left(H_{T}\right)$.
\end{lemma}

Now combining \cref{l: hazan_lemma_5_13} with the above two lemmas, and using $\eta=$ $\frac{D}{\sqrt{2}}$ appropriately, we obtain the theorem.

\end{proof}

We now  can prove \cref{th: final_bound_DAdagrad}.

\paragraph{Proof of \cref{th: final_bound_DAdagrad}.}

\begin{proof}

  We control the terms presented in \cref{eq: general_scheme}. To deal with (A), we exploit \cref{prop: dynamic_adagrad}:

  \begin{align*}
     (A)
     & \leq  \sqrt{2} D\left( 1 + \sqrt{\min_{H \in \mathcal{H}} \sum_{t}\left\|\nabla_{t}\right\|_{H}^{* 2}} \right)
      \end{align*}

    \noindent We now have to deal with (B) and (C) of \cref{eq: general_scheme}.

    \noindent (B) is handled using the strong convexity of $\ell_t$ for any $t:$

    \begin{align*}
         \ell_t(\nu_{t-1}) -  \ell_t(\nu_{t}) & \leq \nabla\ell_t(\nu_{t-1})^\top (\nu_{t-1}-\nu_{t}) - \lambda ||\nu_{t}-\nu_{t-1}||^2 & \\
         & \leq ||\nabla\ell_t(\nu_{t-1})||.|| \nu_{t}-\nu_{t-1}|| - \lambda ||\nu_{t}-\nu_{t-1}||^2 & \text{Cauchy-Schwarz}\\
         & \leq G|| \nu_{t}-\nu_{t-1}|| - \lambda ||\nu_{t}-\nu_{t-1}||^2.
         \intertext{Summing over all $t$ gives us :}
         (B) & \leq GP_T(\nu) - \lambda S_T(\nu).
    \end{align*}

    \noindent To deal with (C), we exploit \cref{l: GD_add_know}.  Indeed, our choice of steps ensure us that at each step $j$: $ \frac{1}{\eta'_j} - \lambda =\lambda\left( j -1\right) = \frac{1}{\eta_{j-1}'}  $.
    We have at each time $t$:
    \begin{align*}
      \ell_t(\nu_{t}) - \ell_t(\xbf_t^*) & \leq \frac{G^2}{K} \sum_{j=1}^K  \eta'_j = \frac{G^2}{\lambda K} \sum_{j=1} \frac{1}{j} \\
      & \leq \frac{G^2 (1+\log(K))}{\lambda K}.
    \end{align*}

    \noindent Finally:

    \begin{align*}
        (C) & \leq T \frac{G^2 (1+\log(K))}{\lambda K} \\
        & = \frac{G^2}{\lambda}(1+\log(T))
    \end{align*}

    \noindent The last line holding because $K = T$.

    \noindent Combining the bounds on (A),(B),(C) concludes the proof.

    \end{proof}

\section{Proofs of probabilistic results of \Cref{sec: convex,sec: additional-OT-algos}}
\label{sec: proof_proba_results}

In this section we use the shortcut $\nabla_t:= \nabla \ell_t(\hat{\xbf}_t)$ and refer to $\xbf_t^*$ as an element of $\operatorname{argmin}_{\xbf \in \mathcal{K}}\ell_t(\xbf)$.
\subsection{The SOCO framework}
\label{sec: SOCO}

In what follows, for a certain filtration $(\mathcal{F}_t)_{t\geq 1}$, we denote by $\mathbb{E}_{t-1}[.]:= \mathbb{E}[.\mid\mathcal{F}_{t-1}]$.
SOCO's framework has been introduced in \cite{wintenberger2021stochastic}. It focuses on a more general notion of regret presented below.

\begin{definition}
    For loss function $\ell_t$, we denote by $(\mathcal{F}_t)_t$ a filtration s.t. $\ell_t$ is $\mathcal{F}_t$-measurable.
    For some predictors $(\hat{\xbf}_t)_{t=1..T}\in\mathcal{K}$ we define the \emph{dynamic averaged regret} with regards to $(\xbf_t)_{t=1..T}\in\mathcal{K}^T$ as follows:

    \[ \text{D-Av-Regret}_T := \sum_{t=1}^T \mathbb{E}_{t-1}[\ell_t(\hat{\xbf}_t)] - \sum_{t=1}^T \mathbb{E}_{t-1}[\ell_t(\xbf_t)].   \]

\end{definition}

\noindent We use SOCO here with the two following assumptions:

\paragraph{(H1)}  The diameter of $\mathcal{K}$ is $D<\infty$ so that $\|x-y\| \leq D, x, y \in \mathcal{K},$ and the functions $\ell_{t}$ are continuously differentiable over $\mathcal{K}$ a.s. and the gradients are bounded by $G<\infty: \sup _{x \in \mathcal{K}}\left\|\nabla \ell_{t}(x)\right\| \leq G$ a.s.,$t \geq 1$

\paragraph{(H2)} The random loss functions $(\ell_{t})$ are stochastically exp-concave i.e. it exists $\alpha>0$ such that, for any $\xbf_1,\xbf_2 \in \mathcal{K}$:
$$
\mathbb{E}_{t-1}[\ell_t(\xbf_2)]\leq \mathbb{E}_{t-1}[\ell_t(\xbf_1)]+\mathbb{E}_{t-1}[\nabla \ell_{t}(\xbf_2)^{T}(\xbf_2-\xbf_1)]-\frac{\alpha}{2} \mathbb{E}_{t-1}\left[\left(\nabla \ell_{t}(\xbf_2)^{T}(\xbf_2-\xbf_1)\right)^{2}\right], \quad x, y \in \mathcal{K}.
$$

\begin{remark}
A $\lambda$-strongly convex function with its gradients bounded by $G$ in absolute value is $\alpha$ stochastically exp-concave with $\alpha= \lambda/G^2$
\end{remark}

Note that Prop 3 of SOCO is valid for dynamic regret:

\begin{lemma}[(Wintenberger 2021, Proposition 3)]
  \label{l: wintenberger_SOCO_2021}
For any decision sequence $(\hat{\xbf}_t)_t\in\mathcal{K}^T,(\xbf_t)_t \in(\mathcal{K}^T)^2$, under $(\mathbf{H 1})$ and $(\mathbf{H} 2)$, with probability $1-\delta$, it holds for any $\beta>0$ and any $T \geq 1$

$$
\begin{aligned}
\sum_{t=1}^{T} \mathbb{E}_{t-1}[\ell_t(\hat{\xbf}_t)]-\sum_{t=1}^{T} \mathbb{E}_{t-1}[\ell_t(\xbf_t)] \leq & \sum_{t=1}^{T} \nabla \ell_{t}\left(\hat{\xbf}_t\right)^{T}\left(\hat{\xbf}_t-\xbf_{t}\right) \\
& +\frac{\beta}{2} \sum_{t=1}^{T}\left(\nabla \ell_{t}\left(\hat{\xbf}_t\right)^{T}\left(\hat{\xbf}_t-\xbf_{t}\right)\right)^{2} +\frac{2}{\beta} \log \left(\delta^{-1}\right) \\
&+\frac{\beta-\alpha}{2} \sum_{t=1}^{T} \mathbb{E}_{t-1}\left[\left(\nabla \ell_{t}\left(\hat{\xbf}_t\right)^{T}\left(\hat{\xbf}_t-\xbf_{t}\right)\right)^{2}\right]
\end{aligned}
$$
\end{lemma}

\subsection{Proof of \cref{th: av-D-OGD}}

Our goal is now to combine this property with our dynamic OGD. To do so, we want to control the quadratic terms in \cref{l: wintenberger_SOCO_2021}. This is the goal of \cref{prop: SOCO_with_Azuma}.

\begin{proposition}
    \label{prop: SOCO_with_Azuma}
    \noindent For any decision sequence $(\hat{\xbf}_t)_t$, any sequence $(\xbf_t)_t$ such that for any $t; (\hat{\xbf}_t, \xbf_t)$ is $\mathcal{F}_{t-1}$- measurable, with probability $1-2\delta$, it holds for any $T \geq 1$

     $$
    \begin{aligned}
    \sum_{t=1}^{T} \mathbb{E}_{t-1}[\ell_t(\hat{\xbf}_t)]-\sum_{t=1}^{T} \mathbb{E}_{t-1}[\ell_t(\xbf_t)] \leq & \sum_{t=1}^{T} \nabla \ell_{t}\left(\hat{\xbf}_t\right)^{T}\left(\hat{\xbf}_t-\xbf_{t}\right) +\left(2(GD)^2 + 6\frac{G^2}{\lambda}\right)\log \left(\delta^{-1}\right)
    \end{aligned}
    $$
\end{proposition}

\begin{proof}

  We define $Y_t=  \nabla \ell_{t}\left(\hat{\xbf}_t\right)^{T}\left(\hat{\xbf}_t-\xbf_{t}\right) $. Remark that $|Y_t|\leq GD$ a.s, we then exploit a corollary of a Poissonian inequality stated in \cite[Eq. (7)]{wintenberger2021stochastic}. With probability $1-\delta$ we have:

  \begin{align*}
    \sum_{t=1}^T Y_t^2 \leq 2\sum_{t=1}^T \mathbb{E}_{t-1}[Y_t^2] + 2(GD)^2\log(1/\delta)
  \end{align*}

  Thus, taking a union bound to make hold this inequality simultaneously with the one of \cref{l: wintenberger_SOCO_2021} and taking $\beta$ such that $3\beta -\alpha =0$ gives us with probability $1 -2\delta$:

  \begin{align*}
    \sum_{t=1}^{T} \mathbb{E}_{t-1}[\ell_t(\hat{\xbf}_t)]-\sum_{t=1}^{T} \mathbb{E}_{t-1}[\ell_t(\xbf_t)] \leq & \sum_{t=1}^{T} \nabla \ell_{t}\left(\hat{\xbf}_t\right)^{T}\left(\hat{\xbf}_t-\xbf_{t}\right) +\left(2(GD)^2 + 6\frac{G^2}{\lambda}\right)\log \left(\delta^{-1}\right)
  \end{align*}

  This concludes the proof.
\end{proof}

\noindent We are now able to prove \cref{th: av-D-OGD}:

\paragraph{Proof of \cref{th: av-D-OGD}.}

\begin{proof}
  We first state that for any $(\hat{\xbf}_t,\xbf_t)$:

  \begin{align*}
      \sum_{t=1}^T \mathbb{E}_{t-1}[\ell_t(\hat{\xbf}_t)] - \sum_{t=1}^T \mathbb{E}_{t-1}[\ell_t(\xbf_t)] & = \sum_{t=1}^T\mathbb{E}_{t-1}\left[ \ell_t(\hat{\xbf}_t) - \ell_t(\xbf_t)   \right]\\
      & \leq \sum_{t=1}^T\mathbb{E}_{t-1}\left[ \ell_t(\hat{\xbf}_t) - \ell_t(\xbf_t^*)   \right] & \text{ with $\xbf_t^* =\operatorname{argmin}_{\xbf\in \mathcal{K}}\ell_t(\xbf)$ } \\
      & = \underbrace{\sum_{t=1}^T\mathbb{E}_{t-1}\left[ \ell_t(\hat{\xbf}_t) - \ell_t(\nu_{t-1})   \right]}_{:=S_1} + \underbrace{\sum_{t=1}^T\mathbb{E}_{t-1}\left[ \ell_t(\nu_{t-1}) - \ell_t(\nu_{t})   \right]}_{:=S_2} \\
      & + \underbrace{\sum_{t=1}^T\mathbb{E}_{t-1}\left[ \ell_t(\nu_{t}) - \ell_t(\xbf_t^*)   \right]}_{:=S_3}
  \end{align*}

  The sum $S_1$ is controlled by applying \cref{prop: SOCO_with_Azuma}.  Then the sum $\sum_{t=1}^T \nabla\ell_t(\hat{\xbf}_t)^T(\hat{\xbf}_t-\nu_{t-1})$ is handled by \cref{prop: D-regret_convex}. We then obtain with our specific choice of steps:

          \begin{align*}
              S_1 & \leq \frac{3}{2}GD\sqrt{T} +\left(2(GD)^2 + \frac{6}{\alpha}\right)\log \left(\delta^{-1}\right) = O(\sqrt{T}) .
          \end{align*}

  \noindent To control the two last sums, we exploit some arguments provided in \cref{th: D_regret_bound_general}. More precisely we use the bounds designed to control the sum (B) and (C) in the \cref{th: D_regret_bound_general}'s' proof. We then have for any $t\geq 0$, by strong convexity of the losses:

\begin{align*}
    \ell_t(\nu_{t-1}) -  \ell_t(\nu_{t}) &  \leq G|| \nu_{t}-\nu_{t-1}||.
    \intertext{Also, our choice of steps, combined with \Cref{l: GD_add_know} gives:}
    \ell_t(\nu_{t}) - \ell_t(\xbf_t^*) & \leq \frac{3}{2\sqrt{K_t}}GD.
  \end{align*}

  \noindent Then, applying our conditional expectations, recalling that $K_t=t$ and summing over $t$ gives us.

    $$S_2  \leq\sum_{t=1}^T \mathbb{E}_{t-1}\left[G\| \nu_{t}-\nu_{t-1}\|  \right],$$

    $$S_3 \leq \frac{3}{2}GD \sum_{t=1}^T \frac{1}{\sqrt{t}}
        \leq 3GD \sqrt{T}.$$

\noindent To conclude the proof, one remarks that if one defines $$ M_T:= \sum_{t=1}^T \mathbb{E}_{t-1}\left[G\| \nu_{t}-\nu_{t-1}\|  \right] - GP_T(\nu) $$

\noindent Then:

\[ S_2 \leq\sum_{t=1}^T \mathbb{E}_{t-1}\left[G\| \nu_{t}-\nu_{t-1}\|  \right] = M_T + GP_T(\nu) \]

\noindent $(M_t)_{t\geq 0}$ is a martingale and furthermore for any $t\geq 0$, $0 \leq \underbrace{G\| \nu_{t}-\nu_{t-1}\|}_{= M_t-M_{t-1}} \leq GD$.

\noindent Thus, applying Azuma-Hoeffding's inequality gives us, with probability $1-\delta$ that $M_T \leq O(\sqrt{T})$

\noindent So with probability $1-\delta$, one has $S_2\leq GP_T(\nu) + O(\sqrt{T})$.

\noindent Applying an union bound on the bounds of $S_1,S_2$ and summing the bound of $S_1,S_2,S_3$ concludes the proof.

\end{proof}

\subsection{Proof of \cref{th: av-OT-ONS}}

\begin{proof}

  We first state that for any $(\hat{\xbf}_t,\xbf_t)$:

  \begin{multline*}
      \sum_{t=1}^T \mathbb{E}_{t-1}[\ell_t(\hat{\xbf}_t)] - \sum_{t=1}^T \mathbb{E}_{t-1}[\ell_t(\xbf_t)]  = \sum_{t=1}^T\mathbb{E}_{t-1}\left[ \ell_t(\hat{\xbf}_t) - \ell_t(\xbf_t)   \right]\\
       \leq \sum_{t=1}^T\mathbb{E}_{t-1}\left[ \ell_t(\hat{\xbf}_t) - \ell_t(\xbf_t^*)   \right]  \text{ with $\xbf_t^* =\operatorname{argmin}_{\xbf\in \mathcal{K}}\ell_t(\xbf)$ } \\
       = \underbrace{\sum_{t=1}^T\mathbb{E}_{t-1}\left[ \ell_t(\hat{\xbf}_t) - \ell_t(\nu_{t})   \right]}_{:=S_1} + \underbrace{\sum_{t=1}^T\mathbb{E}_{t-1}\left[ \ell_t(\nu_{t}) - \ell_t(\xbf_t^*)   \right]}_{:=S_2}
  \end{multline*}

  The sum $S_1$ is controlled by applying \cref{l: wintenberger_SOCO_2021}.  We then obtain with $Y_t= \langle\nabla_t, \hat{\xbf}_{t}- \nu_{t}\rangle$ with probability $1- \delta$  :

  \begin{align*}
      S_1 & \leq \sum_{t=1}^T Y_t + \frac{\beta}{2}\sum_{t=1}^T Y_t^2  + \frac{\beta-\alpha}{2} \mathbb{E}_{t-1}[Y_t^2] + \frac{2}{\beta} \log(1/\delta) .
  \end{align*}

  The first sum is controlled by an intermediary result given in \cref{l: dynamic_ONS}, the second by Cauchy-Schwarz, we then have:

  \begin{align*}
    \sum_{t=1}^T Y_t &  = \sum_{t=1}^T \langle\hat{\xbf}_t-\nabla_t, \hat{\nu}_{t}\rangle + \langle\nabla_t, \hat{\nu}_{t}-\nu_{t}\rangle \\
    & \leq \frac{1}{2 \gamma} \sum_{t=1}^{T} \nabla_{t}^{\top} A_{t}^{-1} \nabla_{t}+\frac{\gamma}{2} \sum_{t=1}^{T}\left(\hat{\xbf}_{t}-\nu_{t-1}\right)^{\top} \nabla_{t}
    \nabla_{t}^{\top}\left(\hat{\xbf}_{t}-\nu_{t-1}\right) + \frac{1}{2\gamma}+ GP_T(\nu)
  \end{align*}

  Recall that, because $\gamma= \frac{1}{2}\min(\frac{1}{GD},\alpha/4)$, $\frac{1}{\gamma}\leq 2\left( \frac{4}{\alpha} + GD   \right)$, one has $\sum_{t=1}^{T} \nabla_{t}^{\top} A_{t}^{-1} \nabla_{t} \leq  2\left( \frac{8}{\alpha} + GD   \right)d\log(T)$.
  Finally, one has:

  \[ \sum_{t=1}^T Y_t \leq  2\left(1+ \frac{8}{\alpha} + GD   \right)d\log(T) + \frac{\alpha}{16}  \sum_{t=1}^{T}  \left(\nabla_{t}^{\top}\left(\hat{\xbf}_{t}-\nu_{t-1}\right) \right) ^2 + GP_T(\nu)   \]

  Plus, remarking that:
\begin{align*}
  \left(\nabla_{t}^{\top}\left(\hat{\xbf}_{t}-\nu_{t-1}\right) \right) ^2  & = \left(\nabla_{t}^{\top}\left(\hat{\xbf}_{t}-\nu_{t}\right)
  + \nabla_{t}^{\top}\left(\nu_{t}-\nu_{t-1}\right) \right) ^2 \leq 2 Y_t^2
  + 2  \left(\nabla_{t}^{\top}\left(\nu_{t}-\nu_{t-1}\right)\right)^2 \\
  & \leq 2 Y_t^2
  + 2  G^2 \| \nu_{t}-\nu_{t-1}\|^2
\end{align*}

Summing on $t$ and reorganising the previous bounds finally gives:

\begin{align*}
  S_1 \leq GP_T(\nu) + 2G^2 S_T(\nu) + \frac{\beta +\alpha/4}{2}\sum_{t=1}^T Y_t^2 + \frac{\beta-\alpha}{2} \mathbb{E}_{t-1}[Y_t^2] + \frac{2}{\beta} \log(1/\delta)  + O(d\log(T))
\end{align*}

Finally, because $|Y_t|\leq GD$ a.s,  we exploit a corollary of a Poissonian inequality stated in \cite[Eq. (7)]{wintenberger2021stochastic}. With probability $1-\delta$ we have:

\begin{align}
  \label{eq: poisson_wint_2021}
  \sum_{t=1}^T Y_t^2 \leq 2\sum_{t=1}^T \mathbb{E}_{t-1}[Y_t^2] + 2(GD)^2\log(1/\delta)
\end{align}

Thus, taking an union bound and $\beta$ such that $3\beta -\alpha/2 =0$ gives us with probability $1 -2\delta$:

\[ S_1 \leq O(d\log(T)) + GP_T(\nu) + G^2S_T(\nu) +\left( \frac{12}{\alpha} + \frac{10\alpha}{24}(GD)^2 \right) \log(1/\delta)   \]

  \noindent Finally, to control $S_2$, we reuse the arguments provided in \cref{th: final_bound_DONS}. More precisely, we use that the step size of \textsc{Construct} allow us to use \cref{l: GD_add_know} to claim that for any $t\geq 0$:
\begin{align*}
    \ell_t(\nu_{t}) - \ell_t(\xbf_t^*) & \leq \frac{G^2}{ K}\sum_{j=1}^K \eta'_j \\
    & \leq \frac{G^2(1+\log(K))}{\lambda K }
  \end{align*}

  \noindent Then, because $K=T$, applying our conditional expectations and summing over $t$ gives us.

    $$S_2 \leq \frac{G^2}{\lambda}(1+\log(T))= O(\log(T)).$$

Summing $S_1$ and $S_2$ concludes the proof.

\end{proof}

\subsection{Proof of \cref{th: av-OT-Adagrad}}

\begin{proof}

  We first state that for any $(\hat{\xbf}_t,\xbf_t)$:

  \begin{multline*}
      \sum_{t=1}^T \mathbb{E}_{t-1}[\ell_t(\hat{\xbf}_t)] - \sum_{t=1}^T \mathbb{E}_{t-1}[\ell_t(\xbf_t)]  = \sum_{t=1}^T\mathbb{E}_{t-1}\left[ \ell_t(\hat{\xbf}_t) - \ell_t(\xbf_t)   \right]\\
       \leq \sum_{t=1}^T\mathbb{E}_{t-1}\left[ \ell_t(\hat{\xbf}_t) - \ell_t(\xbf_t^*)   \right]  \text{ with $\xbf_t^* =\operatorname{argmin}_{\xbf\in \mathcal{K}}\ell_t(\xbf)$ } \\
       = \underbrace{\sum_{t=1}^T\mathbb{E}_{t-1}\left[ \ell_t(\hat{\xbf}_t) - \ell_t(\nu_{t})   \right]}_{:=S_1} + \underbrace{\sum_{t=1}^T\mathbb{E}_{t-1}\left[ \ell_t(\nu_{t}) - \ell_t(\xbf_t^*)   \right]}_{:=S_2}
  \end{multline*}

  The sum $S_1$ is controlled by applying \cref{l: wintenberger_SOCO_2021}.  We then obtain with $Y_t= \langle\nabla_t, \hat{\xbf}_{t}- \nu_{t}\rangle$ with probability $1- \delta$  :

  \begin{align*}
      S_1 & \leq \sum_{t=1}^T Y_t + \frac{\beta}{2}\sum_{t=1}^T Y_t^2  + \frac{\beta-\alpha}{2} \mathbb{E}_{t-1}[Y_t^2] + \frac{2}{\beta} \log(1/\delta) .
  \end{align*}

  The first sum is controlled by an intermediary result given in \cref{prop: dynamic_adagrad}, the second by Cauchy-Schwarz, we then have:

  \begin{align*}
    \sum_{t=1}^T Y_t &  = \sum_{t=1}^T \langle\hat{\xbf}_t-\nabla_t, \hat{\nu}_{t}\rangle + \langle\nabla_t, \hat{\nu}_{t}-\nu_{t}\rangle \\
    & \leq  \sqrt{2} D\left( 1 + \sqrt{\min_{H \in \mathcal{H}} \sum_{t}\left\|\nabla_{t}\right\|_{H}^{* 2}} \right) + GP_T(\nu)
  \end{align*}

Reorganising the previous bounds finally gives:

\begin{align*}
  S_1 \leq GP_T(\nu) + \frac{\beta}{2}\sum_{t=1}^T Y_t^2 + \frac{\beta-\alpha}{2} \mathbb{E}_{t-1}[Y_t^2] + \frac{2}{\beta} \log(1/\delta)
\end{align*}

Finally, because $|Y_t|\leq GD$ a.s,  we exploit a corollary of a Poissonian inequality stated in \cite[Eq. (7)]{wintenberger2021stochastic}. With probability $1-\delta$ we have:

\begin{align}
  \sum_{t=1}^T Y_t^2 \leq 2\sum_{t=1}^T \mathbb{E}_{t-1}[Y_t^2] + 2(GD)^2\log(1/\delta)
\end{align}

Thus, taking an union bound and $\beta$ such that $3\beta -\alpha =0$ gives us with probability $1 -2\delta$:

\[ S_1 \leq \sqrt{2} D\left( 1 + \sqrt{\min_{H \in \mathcal{H}} \sum_{t}\left\|\nabla_{t}\right\|_{H}^{* 2}} \right) + GP_T(\nu) +\left( \frac{2}{\alpha} + \frac{2\alpha}{3}(GD)^2 \right) \log(1/\delta)   \]

\noindent Finally, to control $S_2$, we reuse the arguments provided in \cref{th: final_bound_DONS}. More precisely, we use that the step size of \textsc{Construct} allow us to use \cref{l: GD_add_know} to claim that for any $t\geq 0$:
\begin{align*}
  \ell_t(\nu_{t}) - \ell_t(\xbf_t^*) & \leq \frac{G^2}{ K}\sum_{j=1}^K \eta'_j \\
  & \leq \frac{G^2(1+\log(K))}{\lambda K }
\end{align*}

\noindent Then, because $K=T$, applying our conditional expectations and summing over $t$ gives us.

  $$S_2 \leq \frac{G^2}{\lambda}(1+\log(T))= O(\log(T)).$$

Summing $S_1$ and $S_2$ concludes the proof.

\end{proof}

\end{document}